\pdfoutput=1

\documentclass[11pt]{article}

\newcommand{\highlight}[1]{\textcolor{red}{\textbf{#1}}}
\newcommand{\fcnlp}{\textsc{FcNlp}}

\usepackage{lipsum}  
\usepackage{booktabs}
\usepackage{adjustbox}
\usepackage{url}
\usepackage{amssymb}
\usepackage{amsmath}
\usepackage{multirow}
\usepackage{tabularx}
\usepackage{graphicx}
\usepackage{enumitem}
\usepackage{pifont} 
\usepackage{todonotes}
\usepackage[]{EMNLP2022}

\usepackage{times}
\usepackage{latexsym}

\usepackage[T1]{fontenc}

\usepackage[utf8]{inputenc}

\usepackage{microtype}

\usepackage{inconsolata}

\title{Missing Counter-Evidence Renders NLP Fact-Checking Unrealistic for Misinformation}

\author{
Max Glockner$^{1}$,
Yufang Hou$^{2}$,
Iryna Gurevych$^{1}$ \\
 $^{1}$ Ubiquitous Knowledge Processing Lab (UKP Lab), \\
Department of Computer Science and Hessian Center for AI (hessian.AI), Technical University of Darmstadt  \\
$^{2}$ IBM Research Europe, Ireland\\
\texttt{www.ukp.tu-darmstadt.de},  \texttt{yhou@ie.ibm.com}
}

\begin{document}

\maketitle
\begin{abstract}
Misinformation emerges in times of uncertainty when credible information is limited. This is challenging for NLP-based fact-checking as it relies on counter-evidence, which may not yet be available. Despite increasing interest in automatic fact-checking, it is still unclear if automated approaches can realistically refute harmful real-world misinformation. Here, we contrast and compare NLP fact-checking with how professional fact-checkers combat misinformation in the absence of counter-evidence. In our analysis, we show that, by design, existing NLP task definitions for fact-checking cannot refute misinformation as professional fact-checkers do for the majority of claims. 
We then define two requirements that the evidence in datasets must fulfill 
for realistic fact-checking: 
It must be (1) sufficient to refute the claim and (2) not leaked from existing fact-checking articles. We survey existing fact-checking datasets and find that all of them fail to satisfy both criteria. Finally, we perform experiments to demonstrate that models trained on a large-scale fact-checking dataset rely on leaked evidence, which makes them unsuitable in real-world scenarios.
Taken together, we show that current NLP fact-checking cannot realistically combat real-world misinformation because it depends on unrealistic assumptions about counter-evidence in the data\footnote{Code provided at \url{https://github.com/UKPLab/emnlp2022-missing-counter-evidence}}.
\end{abstract}

\section{Introduction}

According to \newcite{van2022misinformation}, misinformation is ``\emph{false or misleading information masquerading as legitimate news, regardless of intent}''. 
Misinformation is dangerous as it can directly impact human behavior and have harmful real-world consequences such as the Pizzagate shooting~\citep{fisher2016pizzagate}, interfering in the 2016 democratic US election \citep{bovet2019influence}, or the promotion of false COVID-19 cures \citep{aghababaeian2020alcohol}. 
Surging misinformation during the COVID-19 pandemic, coined ``infodemic'' by WHO \citep{zarocostas2020fight}, exemplifies the danger coming from misinformation.
\begin{figure}
\small
    \centering
    \includegraphics[width=0.9\linewidth]{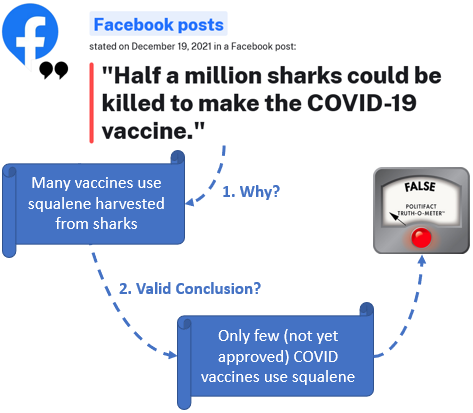}
    \caption{A false claim from PolitiFact. It is unlikely to find counter-evidence. Fact-checkers refute the claim by disproving why it was made.}
    \label{fig:running-example}
\end{figure}
To combat misinformation, journalists from fact-checking organizations (e.g., PolitiFact or Snopes) 
conduct a laborious manual effort to verify claims based on possible harms and their prominence \cite{arnold2020challenges}. 
However, manual fact-checking 
cannot keep pace with the rate at which misinformation is posted and circulated. 
Automatic fact-checking has gained significant attention within the NLP community in recent years, 
with the goal of developing 
tools to assist 
fact-checkers in combating misinformation. For the past few years, NLP researchers have created a wide range of
fact-checking datasets with
claims from fact-checking organization websites 
\citep{vlachos-riedel-2014-fact,wang-2017-liar,augenstein-etal-2019-multifc, hanselowski-etal-2019-richly, ostrowski2020multi, gupta2021xfact, khan-etal-2022-watclaimcheck}. 
 The fundamental goal of fact-checking is, given a \textit{claim} made by a \textit{claimant},
 to find a collection of \textit{evidence} 
   and provide a \textit{verdict} about the claim's veracity based on the evidence. 
 The underlying technique used by fact-checkers, and journalists in general, to assess the veracity of a claim  is called \textit{verification} \citep{silverman2016verification}.
 In a 
 comprehensive survey, \citet{guo2022survey} proposed an NLP fact-checking framework (\fcnlp{}) that aggregates existing (sub)tasks and approaches 
 of automated fact-checking.
\fcnlp{} reflects current research trends on automatic fact-checking in NLP
 and divides the aforementioned process into \emph{evidence retrieval}, \emph{verdict prediction}, and \emph{justification production}.

In this paper, we focus on harmful misinformation claims that satisfied the professional fact-checkers' selection criteria and 
refer to them as \textit{real-world misinformation}.
Our goal is to answer the following research question: \textbf{Can evidence-based NLP fact-checking approaches in \fcnlp{} 
refute novel
real-world
misinformation?}
\fcnlp{} assumes  
a system has access to counter-evidence (e.g., through information retrieval) to refute a claim. 
Consider the false claim ``\emph{Telemundo is an English-language television network}'' from \textsc{Fever}~\citep{thorne-etal-2018-fever}: 
A system following \fcnlp{} must find counter-evidence contradicting the claim (i.e., \emph{Telemundo is a Spanish company}) 
to refute the claim.
This may require more complex reasoning over multiple documents.
We contrast this example to the real-world false claim that ``\emph{Half a million sharks could be killed to make the COVID-19 vaccine}'' (Figure~\ref{fig:running-example}). If true, credible sources would likely report this incident, providing supporting evidence. As it is not, before being fact-checked, there is \textit{no refuting evidence} stating that COVID-19 vaccine production will not kill sharks. Only after 
\textit{guaranteeing} that the claim relies on the false premise of COVID-19 vaccines using squalene (harvested from sharks), it can be refuted. 
After the claim's 
verification, fact-checkers publish reports explaining the verdict and thereby produce counter-evidence. 
Relying on counter-evidence leaked from such reports is unrealistic if a system is to be applied to new claims. 

In this work, we identify
gaps between current research on \fcnlp{} and 
the verification process of professional fact-checkers. 
Via analysis from different perspectives, we argue that the assumption of the existence of counter-evidence in \fcnlp{} is unrealistic and does not reflect real-world requirements. We hope our analysis sheds light on future research directions in automatic fact-checking.
In summary, our major contributions are:
\begin{itemize}
\setlength\itemsep{0.1em}
    \item We identify two criteria from the journalistic verification process, which allow overcoming the reliance on counter-evidence (Section~\ref{sec:human-fact-checking}).
    
    \item We show that \fcnlp{} is incapable of satisfying these criteria, preventing the successful verification of most misinformation claims from the journalistic perspective (Section ~\ref{sec:theoretical-analysis}). 
    
    \item 
    We identify two evidence criteria (\textit{sufficient} \& \textit{unleaked}) for realistic fact-checking. We find that all existing datasets in \fcnlp{} containing real-world misinformation violate at least one criterion (Section~\ref{sec:dataset-overview}) and are hence \textit{unrealistic}.
    
    \item 
    We semi-automatically analyze \textsc{MultiFC}, a large-scale fact-checking dataset to support our findings, and show that models trained on claims from PolitiFact and Snopes (via \textsc{MultiFC}) rely on leaked evidence.
\end{itemize}
\begin{figure}[t]
\small
    \centering
    \includegraphics[width=1.0\linewidth]{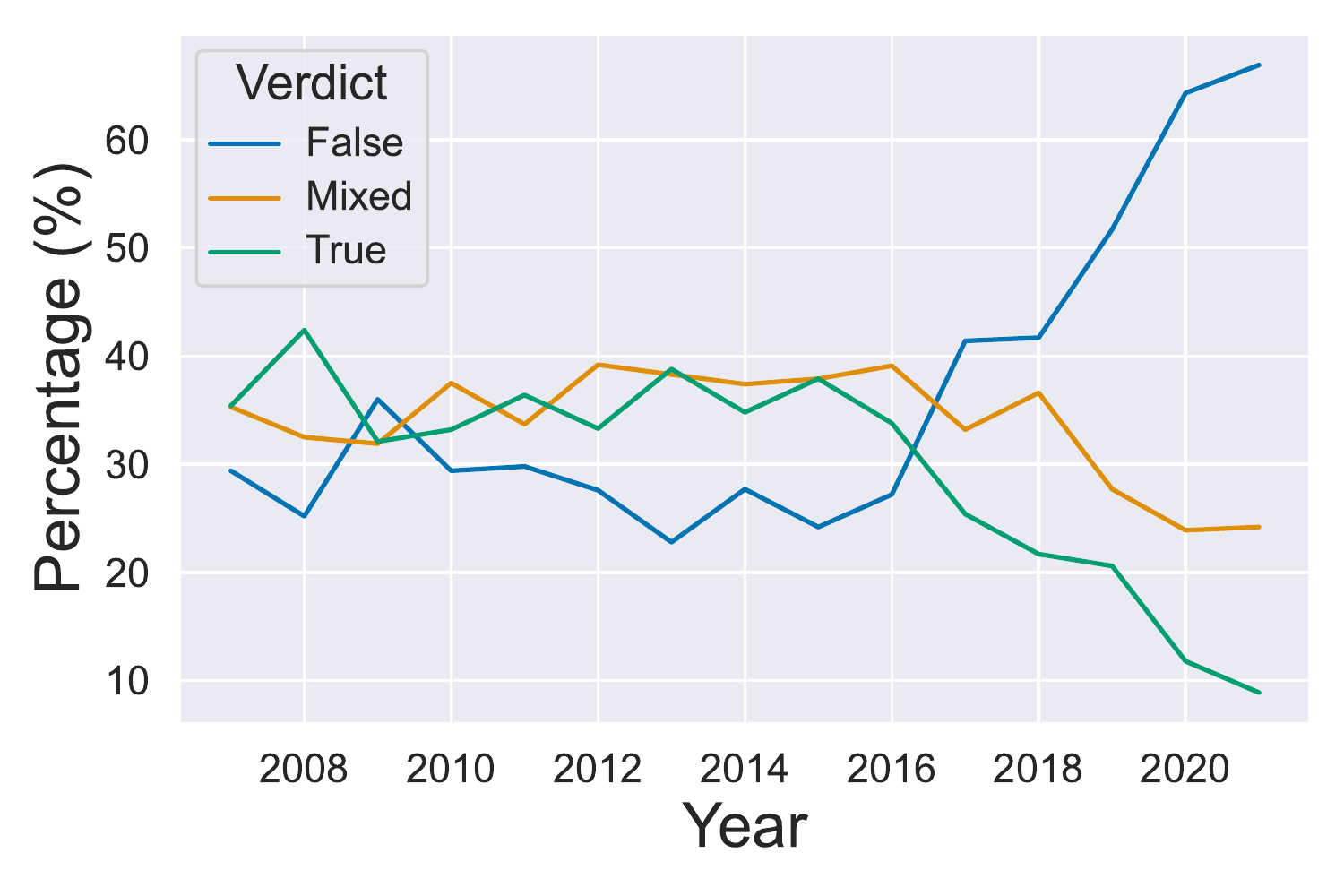}
    \caption{Ratio of verdicts per year (PolitiFact).}
    \label{fig:politifact-counts-relative}
\end{figure}
\section{How Humans Fact-check}
\label{sec:human-fact-checking}

To motivate our distinct focus on \textit{misinformation}, we investigate what claims professional fact-checkers verify.
We crawl 20,274 fact-checked claims from PolitiFact
\footnote{\url{https://www.politifact.com/}}
ranging from 2007--2021. 
Figure~\ref{fig:politifact-counts-relative} shows the ratio of different verdicts
\footnote{We conservatively group verdicts ``pants on fire'' and ``false'' to \textit{False}, ``mostly false'' and ``half true'' to \textit{Mixed} 
and ``mostly true'' and ``true'' to \textit{True}.} 
per year. After 2016,  fact-checkers increasingly select \textit{false} claims as important for fact-checking. In 2021 less than 10\% of the selected claims were correct. 

\begin{table*}[]
\small
    \centering
    \begin{tabularx}{\textwidth}{c X | l}
    \toprule
    & \textbf{Claim} & \textbf{Based Upon}\\
    \toprule
    
    (1) & If you were forced to use a Sharpie to fill out your ballot, that is voter fraud. & false assumption \\
    
    (2) & The Biden administration will begin "spying" on bank and cash app accounts starting 2022. & tax legislation \\
    
    (3) & Barcelona terrorist is cousins with former President Barack Obama. & satire article   \\
    
    (4) & The Democratic health care plan is a government takeover of our health programs. & health care plan\\
    
    (5) & People in Holland protests against of COVID-19 measures. & protests event \\
    
     \bottomrule
     
    \end{tabularx}
    \caption{Example misinformation claims for source guarantee.}
    \label{tab:reference-guarantee-examples}
\end{table*}

Some claims can be refuted via counter-evidence (as required by \fcnlp{}). 
For example, official statistics can contradict the 
false claim about the U.S. that ``\emph{In the 1980s, the lowest income people had the biggest gains}''.
If the evidence makes it impossible for the claim to be true (e.g., because of mutually exclusive statistics) we refer to the evidence as \textit{global counter-evidence}. Global counter-evidence attacks the textual claim itself without relying on reasoning and sources behind it.
In contrast, to refute the claim that ``\emph{COVID-19 vaccines may kill sharks}'' (Figure~\ref{fig:running-example}), fact-checkers did not rely on global counter-evidence specifically proofing that sharks will 
not be killed
to produce COVID-19 vaccines. Neither is it plausible that such  counter-evidence exists. 
Here, the counter-evidence is bound to the claim's underlying (false) reasoning. The claim is only refuted because it follows the false assumption, not because it was disproved.
The absence of global counter-evidence is not an exceptional problem for this specific claim but is common among misinformation: 
Misinformation surges when the high demand for information cannot be met with a sufficient supply of credible answers \citep{silverman2014verification,fullfact2020incident}.
Non-credible and possibly false and harmful information fill these deficits of credible information \citep{golebiewski2019data,shane2020data}. 
The very existence of misinformation often builds on the absence of credible counter-evidence, which in turn, is essential for \fcnlp{}. 

Professional fact-checkers refute misinformation 
even if no global counter-evidence exists, 
e.g., by rebutting underlying assumptions (Figure~\ref{fig:running-example}). 
Table~\ref{tab:reference-guarantee-examples} shows a few false claims built on top of various resources: (1)
relies on a false assumption that sharpies invalidate election ballots, (2 \& 4) misinterpret 
 official documents or laws,
(3) is based on non-credible sources,
 and (5) changes a topic of a specific event from ``\emph{gas extraction}'' to ``\emph{COVID-19 measures}''.
Fact-checkers use the reasoning \textit{for} the claim to consider 
evidence that is, or refers to, the claimant's source:
the original tax legislation (2), or alternate (correct) descriptions of protests against gas extraction (5).
Here, the content of the evidence alone is often insufficient.
The assertion that the claimant's source and the used counter-evidence are identical, 
or refer to the same event 
is crucial 
to refute the claim: 
Claim (2) is refuted because the tax legislation it relies upon does not support the ``spying'' claim. However, the document does not specifically refute the claim, and without knowing that the claimant relied on it, it becomes useless as counter-evidence.
Similarly, the correct narrative of protests against gas extraction is only mutually exclusive to the false claim (5) of protests against COVID-19 measures when assuring both refer to the identical incident.
For similar reasons, the co-reference assumption is critical to the task definition of SNLI~\citep{bowman-etal-2015-large}.
After this assertion, mutual exclusiveness is not required to refute the claim: It is sufficient if the claim is not entailed (i.e. incorrectly derived or relies on unverifiable speculations) or based on invalid sources (such as satire) to refute it. 
Based on these observations we identify two criteria to refute claims if no global-counter evidence exists. We validate their relevance in Section~\ref{sec:theoretical-analysis}:
\begin{itemize}
\setlength\itemsep{0.1em}
    \item \textbf{\textit{Source Guarantee:}} The guarantee that identified evidence either constitutes or refers to the claimant's reason for the claim. 

    \item \textbf{\textit{Context Availability:}} 
    We broadly consider context as the claim's original environment, which allows us to unambiguously comprehend the claim, and trace the claim and its sources across multiple platforms if required.
    It is a logical precondition for the source guarantee. 
\end{itemize}
Both criteria are challenging for computers but naturally satisfied by human fact-checkers. 
\citet{buttry2014verification} defines the question \textit{``How do you know that?”} to be at the heart of verification.
After selecting a claim, finding provenance and sourcing are the first steps in journalistic verification.
Provenance provides crucial information about context and motivation \citep{urbani2020verifying}.
Journalists must then identify solid sources to compare the claim with \citep{silverman2014verification,borel2016chicago}. Ideally, the claimant provides sources, which must be included and assessed in the verification process. 
During verification, journalists rely, if possible, on relevant primary sources, such as uninterpreted and original legislation documents (for claim 2, Table~\ref{tab:reference-guarantee-examples}).
Fact-checking organisations see sourcing as one of the most important parts of their work \citep{arnold2020challenges}. 

\section{Can \fcnlp{} Help Human Verification?}
\label{sec:theoretical-analysis}
In this section, we first analyze human verification strategies based on an analysis of 100 misinformation claims. We then contrast human verification strategies with \fcnlp{}. 

\subsection{Human Verification Strategies}

We manually analyze 100 misinformation claims\footnote{Claims are from the following categories: ``\emph{pants on fire}'', ``\emph{false}'' and ``\emph{mostly false}''.} 
from two well-known fact-checking organizations: PolitiFact and Snopes.
We 
randomly choose 50 misinformation claims from each website which contains 25 claims from \textsc{MultiFC} (a large NLP fact-checking dataset with real-world claims before 2019) and 25 claims from 2020/2021. 
We extract the URL for each claim and 
analyze its verification strategy 
based on the entire fact-checking article.
Claims that require the identification of scam webpages, imposter messages, or multi-modal reasoning\footnote{
If a claim can be expressed in text and verified without  multi-modal reasoning we 
consider the verbalized variant of the claim and
do not discard it.}
such as detecting misrepresented, miscaptioned or manipulated images \citep{zlatkova-etal-2019-fact} were marked as not applicable to \fcnlp{} by nature.
In the first round of analysis, we assess whether humans relied on the \textit{source guarantee} to refute the claim. 
Each claim (and its verification) is unique and 
can be refuted using different strategies.
In the second round of analysis we identify the primary strategy to refute the claim and verify that it is 
based on the source guarantee.
This led us to 
identify 4 primary human-verification strategies: 
\begin{table}[]
\small
    \centering
    \begin{tabular}{c l | c c c r}
    \toprule
    \textbf{Src.} & \textbf{Strategy} & \textbf{\textsc{MultiFC}} & \textbf{20/21} & \textbf{All} & \textbf{\%}\\
    \midrule

    yes & LCE & 19 & 16 & 35 & 46.7\\
    yes & NCS & 9 & 5 & 14 & 18.7\\
    
    \midrule
    no   & GCE & 10 & 10 & 20 & 26.7\\
    no  & NEA & 1 & 4 & 5 & 6.7\\
    no  & other & 0 & 1 & 1 & 1.3\\
    \midrule
        \textbf{yes} & \textbf{\textit{all}} & \textbf{28} & \textbf{21} & \textbf{49} & \textbf{65.3} \\
    \textbf{no} & \textbf{\textit{all}} & \textbf{11} & \textbf{15} & \textbf{26} & \textbf{34.7} \\
    \textbf{\textit{all}} & \textbf{\textit{all}} & \textbf{39} & \textbf{36} & \textbf{75} & 100.0\\

     \bottomrule
     
    \end{tabular}
    \caption{
    Strategies used to refute 75 of 100 misinformation claims with and without source guarantee (\textbf{Src.}).
    }
    \label{tab:guarantee-analysis-results}
\end{table} 

\begin{enumerate}
\setlength\itemsep{0.1em}
    \item \textit{Global counter-evidence (GCE):} Counter-evidence via arbitrarily complex reasoning but without the source guarantee.
    
    \item \textit{Local counter-evidence (LCE):} Evidence requires the source guarantee to refute the (reasoning behind) the claim.
    
    \item \textit{Non-credible source (NCS):} Evidence requires the source guarantee to refute the claim based on non-credible sources (e.g. satire).
    
    \item \textit{No evidence assertion (NEA):} The claim is refuted as no (trusted) evidence supports it.

\end{enumerate}
We discard 25 non-applicable claims and show the results
of the remaining 75 claims in Table~\ref{tab:guarantee-analysis-results}. Please refer to Appendix \ref{appendix:human-misinformation:examples} for more analysis details and examples. 
In some cases, the selection of one strategy is ambiguous if multiple strategies are applied.
In a pilot study to analyze human verification strategies, two co-authors agreed on 9/10 applicable misinformation claims.
In general, about two-thirds of the claims were refuted by relying on the source guarantee. In 20 cases fact-checkers
refuted the claim by finding global counter-evidence.
In one case (\textit{other}), fact-checkers relied entirely on expert statements.
In general, experts supported the fact-checkers in identifying and discussing evidence, or strengthened their argument via statements
but did not 
affect the 
underlying verification
strategy. 

\subsection{NLP Fact Verification}
\label{sec:theoretical-analysis:nlp}
\paragraph{Focusing on evidence-based approaches.} Approaches in \fcnlp{} estimate the claim's veracity based on surface cues within the claim \citep{rashkin-etal-2017-truth,patwa2020fighting}, 
assisted with metadata \citep{wang-2017-liar,cui2020coaid,yichuan2020toward,dadgar2021checkovid}, or using evidence documents. 
Here, the system  
uses 
the stance of the evidence towards the claim to predict the verdict. 
Verdict labels are often non-binary and include a neutral stance \citep{thorne-etal-2018-fever}, or fine-grained veracity labels from fact-checking organizations \citep{augenstein-etal-2019-multifc}. 
Evidence-based approaches either rely on unverified documents or user comments \citep{ferreira-vlachos-2016-emergent,zubiaga-etal-2016-stance,pomerleau2018fake}, or assume access to a presumed trusted knowledge base such as Wikipedia \citep{thorne-etal-2018-fever},  scientific publications \citep{wadden-etal-2020-fact}, or search engine results \citep{augenstein-etal-2019-multifc}. 
In this paper, we focus on trusted evidence-based  verification approaches which 
can deal with the truth changing over time \citep{schuster-etal-2019-towards}. 
More importantly, they are the most representative of professional fact verification. 
Effectively debunking misinformation requires stating the corrected fact and explaining the myth's fallacy \citep{lewandowsky2020debunking}, both of which require trusted evidence.

\paragraph{Global counter-evidence assumption in \fcnlp{}.}
In \fcnlp{}, evidence retrieval-based approaches assume that the semantic content of a claim 
is sufficient to find relevant (counter-) evidence in a trusted knowledge base \citep{thorne-etal-2018-fever, jiang-etal-2020-hover, wadden-etal-2020-fact, aly2021feverous}.
 This becomes problematic for misinformation 
 that requires the source guarantee to refute the claim. 
 By nature, in this case, the claim and evidence content are distinct and not entailing. Content cannot assert that two different narratives describe the same protests (e.g., Claim 5 in Table~\ref{tab:reference-guarantee-examples}), or that a non-entailing fact (squalene is harvested from sharks) serves as a basis for the false claim (e.g., Figure~\ref{fig:running-example}).
The consequence is a circular reasoning problem: Knowing that a claim is false is a precondition to establishing the
source guarantee, which in turn is needed to refute the claim.
To escape this cycle,
one must (a) provide the source guarantee by other means 
than content
(e.g., context), or (b) find evidence that refutes the claim without the source guarantee (global counter-evidence). 
By relying only on the content of the claim, \fcnlp{} cannot provide the source guarantee and is limited to global counter-evidence, which only accounts for 20\% of misinformation claims analyzed in the previous section.

\paragraph{Current \fcnlp{} fails to provide source guarantees.}
We note that providing the source guarantee goes beyond entity disambiguation, as required
in \textsc{Fever}~\citep{thorne-etal-2018-fever}. 
The self-contained context within claims in \textsc{Fever} is typically sufficient to disambiguate named entities if required.\footnote{In the claim ``\emph{Poseidon grossed \$181,674,817 at the worldwide box office on a budget of \$160 million}'' it is clear that ``Poseidon'' refers to the \textit{film}, not an ancient god. (\textsc{Fever})}
After disambiguation, the 
retrieved evidence serves as global counter-evidence.

Recent approaches further 
add context snippets from Wikipedia \citep{sathe-etal-2020-automated} or dialogues \citep{gupta2021dialfact}
to resolve ambiguities 
and cannot provide the source guarantee to break the circular reasoning problem.
These snippets differ from the context used by professional fact-checkers who often need to trace claims and their sources across different platforms. 
Recently, \citet{thorne2021evidence} annotate more realistic claims w.r.t. multiple evidence passages. They found supporting and refuting passages for the same claim, which prevents the prediction of an overall verdict. 
Some works collect evidence for the respective claims by identifying scenarios where the \textit{claimant's source} is naturally provided:  such as a strictly moderated forum \citep{saakyan2021covidfact}, scientific publications \citep{wadden-etal-2020-fact}, or Wikipedia references \citep{sathe-etal-2020-automated}. 
However, such source evidence 
is only collected for true claims. Adhering to the global counter-evidence assumptions of previous work, false claims in these works are generated artificially and do not reflect real-world misinformation.

\subsection{Human and NLP Comparison}
Our analysis (Table~\ref{tab:guarantee-analysis-results}) finds fact-checkers only refuted 26\% of false claims with global counter-evidence. In all other cases, fact-checkers 
relied on source guarantees (LCE, NCS) or
asserted that no supporting evidence exists (NEA).
The verification strategy is not evident given the claim alone but dependent on existing evidence.
The claim that ``\emph{President Barack Obama’s policies have forced many parts of the country to experience rolling blackouts}'' 
is refuted via global counter-evidence (that rolling blackouts had natural causes). The claim that ``\emph{90\% of rural women and 55\% of all women are illiterate in Morocco}'' seems
verifiable 
via
official statistics. 
Yet, no comparable statistics exist and the claim is refuted due to relying on a decade-old USAID request report.

\newcommand{\tableno}{\textit{no}}
\newcommand{\tableyes}{\checkmark}

\begin{table*}[]
\small
    \centering
    \begin{tabular}{c l|c c | c c | c}
    \toprule
    & & \multicolumn{2}{c|}{\textit{Claims}} & \multicolumn{2}{c|}{\textit{Evidence}} & \textbf{Ev.}\\
    & \textbf{Dataset} & \textbf{Source} & \textbf{False Claims} & \textbf{Unleaked} & \textbf{Sufficient} & \textbf{Ann.}\\
    \toprule
    
    1& \textsc{SciFact} \citep{wadden-etal-2020-fact} & Scientific & \textit{generated} & n/a & \tableyes{}  & \tableyes{}\\
    2& \textsc{Covid-Fact} \citep{saakyan2021covidfact} & Reddit & \textit{generated} & n/a & \tableyes{}  & \tableyes{}\\
    3& \textsc{WikiFactCheck} \citep{sathe-etal-2020-automated} & Wikipedia & \textit{generated} & n/a & \tableyes{} &  \tableyes{}\\ 
     4&  \textsc{Fm2} \citep{eisenschlos-etal-2021-fool} &  Game & \textit{generated} & n/a & \tableyes{} & \tableyes{} \\
    5&\citet{thorne2021evidence} & User Queries & \textit{paraphrased} & n/a & \tableyes{} & \tableyes{}\\ 
    6&\textsc{FaVIQ} \citep{park2021faviq} & User Queries & \textit{paraphrased} & n/a & \tableyes{} & \tableno{}\\
    
    \midrule
    
    7&\textsc{LiarPlus} \citep{wang-2017-liar,alhindi-etal-2018-evidence} & FC Article & \textit{real-world} & \tableno{} & \tableyes{}  & \tableyes{}\\ 
    8&\textsc{PolitiHop} \citep{ostrowski2020multi} & FC Article & \textit{real-world} & \tableno{} & \tableyes{}  & \tableyes{}\\
    
        \midrule
        
    9&\textsc{ClimateFever} \citep{diggelmann2020climatefever}  & Web & \textit{real-world} & \tableyes{} & \tableno{}  & \tableyes{}\\
    10&\textsc{HealthVer}~\citep{sarrouti-etal-2021-evidence-based} & Web & \textit{real-world} & \tableyes{} & \tableno{}  & \tableyes{}\\
    
    \midrule
    
    11&\textsc{UKP-Snopes} \citep{hanselowski-etal-2019-richly} & FC Article & \textit{real-world} & \tableyes{} & \tableno{} & \tableyes{}\\ 
    12&\textsc{PubHealth} \citep{kotonya-toni-2020-explainable} & FC Article & \textit{real-world} & \tableyes{} & \tableno{}  & \tableno{}\\
    13&\textsc{WatClaimCheck} \citep{khan-etal-2022-watclaimcheck} & FC Article & \textit{real-world} &  \tableyes{} & \tableno{}  & \tableno{}\\

    \midrule    
    
    14&\citet{baly-etal-2018-integrating} & FC Article & \textit{real-world} & \tableno{} & \tableno{} & \tableyes{}\\
    15&\textsc{MultiFC} \citep{augenstein-etal-2019-multifc} & FC Article  & \textit{real-world} & \tableno{} & \tableno{}  & \tableno{}\\
    16&\textsc{X-Fact} \citep{gupta2021xfact} & FC Article & \textit{real-world} & \tableno{} & \tableno{}  & \tableno{}\\
    
     \bottomrule
     
    \end{tabular}
    \caption{Overview of NLP fact-checking datasets as realistic test-beds to combat real-world misinformation. We indicate whether humans annotated the stance between claim and evidence (\textbf{Ev. Ann.})}
    \label{tab:overview-fact-checking-datasets}
\end{table*}
We further analyze claims refuted via global counter-evidence, that \fcnlp{}, in theory, can refute.
Some claims only require shallow reasoning as directly contradicting evidence naturally exists: 
A transcript of an interview in which Ron DeSantis was asked about the coronavirus can easily refute the claim ``\emph{Ron DeSantis was never asked about coronavirus}''. 
Another case is when information about the claim's veracity already exists, e.g.,  because those affected by the myth already corrected the claim.
Most claims require complex reasoning like legal text understanding or comparing and deriving statistics. 
Some
claims 
require the definition of some terms first, to make them verifiable.
Collecting all required global counter-evidence often requires aggregating and comparing different information, possibly under time constraints. 
Consider the false claim that ``\emph{Illegal immigration wasn’t a subject that was on anybody’s mind until [Trump] brought it up at [his] announcement}'': 
To refute this claim, one must first determine when Trump announced his run for the presidency, then
count and compare how often ``illegal immigration'' was mentioned before and after the announcement.\footnote{We assume disambiguation requires no source guarantees, and that basic context (date, location, claimant) is known.} 
\section{NLP Fact-Checking Datasets}
\label{sec:dataset-overview}
Based on our observations in Section~\ref{sec:theoretical-analysis} and \fcnlp{}'s reliance on global counter-evidence, we hypothesize that evidence in existing fact-checking datasets does not fully satisfy real-world demands.
We, hence, investigate how \fcnlp{}'s assumptions affect 
fact-checking datasets and if they constitute realistic test beds for real-world misinformation. 
For real-world scenarios, 
datasets must contain real-world misinformation claims and realistic counter-evidence. For evidence,
we define the following two requirements: 
\begin{itemize}
\setlength\itemsep{0.1em}
    \item \textit{Sufficient:} Evidence must be sufficient to justify the verdict from a human perspective. 
    \item \textit{Unleaked:} Evidence must not contain information that only existed after the claim was verified.
\end{itemize}
The issue of leaked evidence was also mentioned very recently by
\citet{khan-etal-2022-watclaimcheck}. Unlike us, they did not comprehensively analyze existing datasets, evaluate the impact on trained systems (Section~\ref{sec:multifc-case-study}), or consider the complementary criterion of \textit{sufficient} (counter-) evidence.
Relying on leaked evidence is related to the important yet different task of detecting already-verified claims \citep{shaar-etal-2020-known}, but is unrealistic for novel claims.

We survey NLP fact-checking datasets with natural input claims\footnote{See the survey from \citet{guo2022survey} for datasets with natural and artificial claims.} that assume
access to trusted evidence.
Table~\ref{tab:overview-fact-checking-datasets} summarizes our 
survey results.
Datasets 1-6 
contain no real-world misinformation:
False claims are derived from true real-world claims (1-3) or within a gamified setting (4), ensuring that counter-evidence exists. Other works (5 \& 6) reformulate real-world user queries, which are linked to Wikipedia articles as (counter-) evidence. 

\begin{table*}[t]
\small
    \centering
    \begin{tabularx}{\textwidth}{l X}
    \toprule
    \textbf{Detected via}  & \textbf{Claim \& Evidence} \\
    \midrule

    phrase \&  & \textbf{Claim}: \textit{Google Earth Finds SOS From Woman Stranded on Deserted Island} \\
    snippet URL & \textbf{Evidence Snippet:} The Truth: The story is a \highlight{hoax}. ... GOOGLE EARTH FINDS WOMAN TRAPPED  ON DESERTED ISLAND FOR 7 YEARS ... other end “How did you find me” to  which they replied “Some kid from Minnesota found your SOS sign on Google  Earth”; From \highlight{\textit{Truth Or Fiction}}\\
    \midrule
    
    phrase &   \textbf{Claim}: \textit{Country music singer Merle Haggard left his entire estate to an LGBT group.} \\
    & \textbf{Evidence Snippet:} Discover ideas about Country Singers. \highlight{Fake news} reports that recently-deceased  country music legend Merle Haggard left his entire estate to an LGBT group; From \textit{Pinterest}\\
     \bottomrule
     
    \end{tabularx}
    \caption{Examples from \textsc{MultiFC} of leaked evidence detected via the \textit{snippet URL} (linking to a fact-checking article) or a \textbf{phrase} of the evidence snippet.
}
    \label{tab:examples-leaking-evidence-multifc}
\end{table*}
We find that no dataset with real-world misinformation (7-16) satisfies both evidence criteria. We identify four 
categories: 
First, datasets that consider (parts of) a fact-checking article as evidence contain sufficient, yet leaked evidence (7 \& 8).
Second, annotators estimate claim veracity based on
evidence such as Wikipedia or scientific publications. The authors 
find that 
evidence often only covers parts of these realistic, complex claims, which yield low annotator agreement (9), or a weakened task definition for stances (10). 

Third is to rely on the same evidence as fact-checkers,
termed \textit{premise evidence} by \citet{khan-etal-2022-watclaimcheck}. 
Here, only \textsc{UKP-Snopes} (11) provides evidence annotations. \citet{hanselowski-etal-2019-richly} collect and annotate original evidence snippets 
from the fact-checking article. 
They find the stance often conflicts with the verdict: 
Though most claims are false, the majority of evidence is supporting. In 45.5\% of cases, annotators found no stance for the professionally selected evidence snippets
even though professional fact-checkers considered these snippets important to be included in the article.
Due to conflicting and unexplained evidence snippets, we rate this \textit{insufficient} to predict the correct verdict. 
The human verification process (Section~\ref{sec:human-fact-checking}) 
guides the creation
of the fact-checking article and
can serve as a possible explanation for these problems.
Articles link to the claim's context and possibly other similar claims (likely \textit{supporting}). Often (e.g. during COVID-19 \citep{brennen2020types}), claims are not completely fabricated.
Fact-checkers identify documents and their interdependence when investigating the claimant's reasoning for the claim
(likely not \textit{refuting}).
Documents used to disprove the claimant's reasoning may have no or little relevance to the original claim (as in Figure~\ref{fig:running-example}).
Each step is non-trivial and may rely on numerous documents (or expert statements). Relying on 
premise evidence 
without considering the verification process and \textit{how} these documents relate, is insufficient. Both other datasets (12 \& 13) in this category provide no annotations and are limited to freely available evidence documents (as opposed to paywalled web pages or e-mails). 

Fourth is using a search engine during dataset construction to expand the accessible knowledge.
Even when excluding search results that point to the claim's fact-checking article, leaked evidence persists: Different organizations may verify the same claims, or disseminate the fact-checkers verification.
Only \citet{baly-etal-2018-integrating} provide stance annotation for Arabic claim and evidence pairs. 
For false claims, they found that only a few documents disagree, and more agree,
with the claim.
A possible explanation is that misinformation often emerges 
when
trusted information or counter-evidence is scarce. 
Fact-checking articles fill this deficit.
Partially excluding them during dataset generation reduces the found counter-evidence. Lacking counter-evidence is not a problem of the dataset generation, but the underlying nature of misinformation, and should be considered by the task definition.
We rate evidence 
in this category 
(14-16) leaked and insufficient, and 
back it up in Section~\ref{sec:multifc-case-study}.

\section{A Case Study of Leaked Evidence}
\label{sec:multifc-case-study}
We view \textsc{MultiFC}~\citep{augenstein-etal-2019-multifc}, the largest dataset of its group, as an instantiation of \fcnlp{} applied to the real world: 
It contains real-world claims and professionally assessed verdicts as labels from 26 fact-checking organizations (like Snopes or PolitiFact). 
The authors use the Google search engine to expand evidence retrieval to the real world during dataset construction.
We abstract from the fact that \textsc{MultiFC} only provides incomplete evidence snippets and consider (if possible) the underlying article in its entirety.

\subsection{Quantification of Leaked Evidence}
We focus our analysis on 16,244 misinformation claims 
that we identify via misinformation labels (listed in Appendix~\ref{app:multifc-analysis:misinformation}).
To quantify how many claims in \textsc{MultiFC} contain leaked evidence, we consider all evidence snippets stemming from a fact-checking article, or discussing the veracity of a claim, as leaked. 
Table~\ref{tab:examples-leaking-evidence-multifc} shows examples of leaked evidence that strongly indicates the claim's verdict. The first snippet comes directly from a fact-checking organization (Truth Or Fiction\footnote{\url{https://www.truthorfiction.com/}}). Only identifying leaked evidence that directly comes from fact-checking organizations is insufficient: After the publication of the verification report, its content is disseminated via other publishers (such as Pinterest in the second example). 
We identify leaked evidence snippets 
using patterns for their source URLs or contained phrases. A complete list of all used patterns is given in Appendix~\ref{app:multifc-analysis:misinformation:automatic-leak-detection}).
This requires the 
evidence to be relevant.
Irrelevant articles are insufficient, albeit not leaking. 
To this end,
we manually analyze 100 claims with 230 automatically found leaking evidence snippets. We confirm that 83.9\% of the snippets are leaked (details in Appendix~\ref{appendix:multifc-analysis:leaked-guidelines}).
97/100 of the selected claims contain at least one leaked evidence snippet.

\begin{table}[]
\small
    \centering
    \begin{tabular}{l|c c }
    \toprule
    \textbf{Leaked} & \textbf{Claims (Number)} & \textbf{Claims (\%)} \\
     \toprule
    Url & 8,999 & 55.6\% \\
    Phrase & 9,656 & 59.7\% \\
    Url or Phrase & 11,267 & 69.7\% \\
    \bottomrule
    \end{tabular}
    \caption{Absolute and relative number of automatically identified leaked evidence of \textsc{MultiFC} misinformation.
    }
    \label{tab:multifc-leakage}
\end{table}
\begin{table}[]
\small
    \centering
    \begin{tabular}{l|ccc}
    \toprule
    & \multicolumn{3}{c}{\textit{Claim has leaked evidence}} \\
    \textbf{Categories} & \textbf{All}  & \textbf{Leaked} & \textbf{Unleaked}\\
     \toprule
     Any & 100& 32 & 68\\
     No Stance & 37 &  0 & 37\\
     
     \midrule
     
     \textbf{No Refuting} & 63 & 0 & 63\\
     \textbf{Original \& Refuting}  & 15 & 10 & 5\\

    \bottomrule
    
    \end{tabular}
    \caption{Manual analysis of 100 claims without automatically identified leaked evidence.}
    \label{tab:multifc-unleaked}
\end{table}

Table~\ref{tab:multifc-leakage} lists the number of claims with leaked evidence identified by the pattern-based approach.
It detected leaked evidence for 69.7\% of misinformation claims. 
In addition, we manually analyze evidence of 100 misinformation claims 
for which this approach found no leaked evidence. Misinformation verification 
often
requires 
multiple
evidence documents, rendering a single sufficient evidence snippet unrealistic. 
We follow \citet{sarrouti-etal-2021-evidence-based} and test if a snippet supports or refutes parts of the claim. 
Table~\ref{tab:multifc-unleaked} shows that approximately one-third of the claims contain further leaked evidence. 15 claims have unleaked refuting evidence. In 10 cases this evidence is overshadowed via leaked evidence for the same claim. Most analyzed claims only have non-refuting evidence.
Similar to \citet{baly-etal-2018-integrating}, we found supporting evidence for 40 misinformation claims;
for 35 of these claims, the evidence was misinformation and thus supported the claim; for the remaining five claims, the claim became accurate, and the evidence became available at a date later than the claim's creation.

\begin{table}[]
\small
    \centering
    \begin{tabular}{l| c c c c}
    \toprule
    \textbf{Input} & \textbf{All}  & \textbf{Leaked} & \textbf{Unleaked} & \textbf{$\Delta$}\\
     \toprule
     
    \multicolumn{5}{c}{\textit{Snopes}}\\
    \midrule
    \textbf{Samples} & 1,014 & 482 & 532& \\
    \midrule
    Sn.-Text & 29.4/60.4 & 26.3/66.3 & 30.1/55.0 & +11.3\\
    Sn.-Title & 27.3/57.8 & 23.8/64.6 & 28.2/51.5 & +13.1\\
    Snippets & 30.5/60.5 & 28.7/67.6 & 30.2/53.7 & \textbf{+13.9}\\
    Full & \textbf{32.7}/\textbf{62.7} & \textbf{30.6}/\textbf{68.7} & \textbf{33.0}/\textbf{57.2} & +11.5\\
     
     \midrule
     \multicolumn{5}{c}{\textit{PolitiFact}}\\
     \midrule
     \textbf{Samples} & 2,717 & 2,111 & 606&\\
     \midrule
    Sn.-Text & 35.5/34.5 & 38.0/37.2 & 24.1/24.5 & +12.7\\
    Sn.-Title & 48.0/47.4 & 55.1/54.6 & 21.1/21.6 & +33.0\\
    Snippets & 52.0/51.3 & \textbf{59.7}/59.2  & 22.1/23.0  & \textbf{+36.2}\\
    Full & \textbf{52.6}/\textbf{51.9} & \textbf{59.7}/\textbf{59.3} & \textbf{25.6}/\textbf{25.9} & +33.4\\

    \bottomrule
    \end{tabular}
    \caption{
    F1-macro/micro scores and difference in F1-micro ($\Delta$) averaged over 3 runs.
    Inputs are: only snippet texts, titles, entire snippets or claim and snippets (full).
    }
    \label{tab:multifc-experiments}
\end{table}

\subsection{Impact on Trained Systems}
\citet{hansen2021automatic} found that models in \textsc{MultiFC} can predict the correct verdict based on the evidence snippets alone.
To test if
leaked evidence can serve as an explanation, 
we fine-tune BERT~\citep{devlin2018bert} (\textit{bert-base-uncased}) to predict the veracity label of a claim given the evidence snippets with and without a claim. As input to BERT, we separate the claim (when used) from the evidence snippets using a \textsc{[SEP]} token and predict the veracity label based on a linear layer on top of a preceding \textsc{[CLS]} token (Training details in  Appendix~\ref{app:multifc-analysis:experiment-setup}.).
When each evidence snippet is represented via its content only
this 
performs on par with the specialized model introduced by \citet{hansen2021automatic}. We additionally find that the snippet's title carries much signal, 
and adding it to the input improves the overall performance on PolitiFact. Snippets are concatenated (separated by ``;'' in the order provided by \textsc{MultiFC} and truncated after 512 tokens.
We experiment on the train-, dev- and test-splits \citet{hansen2021automatic} extracted from \textsc{MultiFC} on claims from Snopes and PolitiFact.
We test four types of input: only evidence (only title, only text, both),
or the complete sample of claim and evidence.
For the evaluation (Table~\ref{tab:multifc-experiments}), we split the test data based on whether a claim contains automatically identified leaked evidence. 

On Snopes, the macro-F1 
is higher on the unleaked than on the leaked subset. 
Upon closer inspection, we find that the label distribution on Snopes is heavily skewed towards ``false'', which worsens on the leaked subset. Models seem to rely on patterns of leaked evidence to predict the majority label ``false'' (see Appendix~\ref{app:multifc-analysis:performance-by-label}).
On the leaked subset, this comes at the cost of incorrect predictions for all other labels, yielding a lower F1-macro.
On the larger PolitiFact subset, labels are not much skewed towards a single majority label. Across all experiments, 
the performance gap signals the reliance on leaked evidence. 
We confirm the impact of leaked evidence for both datasets by evaluating the model on the \textit{same} instances with leaked \textit{or} unleaked evidence,
to avoid the different label distribution distorting the results (Appendix~\ref{app:multifc-analysis:prformance-by-only-leaked-unleaked}). 

\section{Related Work}
\label{sec:future-directions}
\paragraph{Combat Misinformation After Its Verification.}
The identified limitations of the previous studies on NLP fact-checking datasets described in Section~\ref{sec:dataset-overview} do not devalue the surveyed datasets and we view them as highly important and useful contributions. These limitations are tied to our specific research question to refute \textit{novel} real-world misinformation. We strongly build on these previous works and view them as crucial starting points to fact-check real-world misinformation. Existing fact-checking articles are highly valuable and automatic methods should utilize them to detect and combat misinformation. Automatic methods specifically using these resources detect misinformation by matching claims with known misconceptions \citep{hossain-etal-2020-covidlies,weinzierl2022vaccinelies} or already verified claims \citep{vo-lee-2020-facts, shaar-etal-2020-known,martin2021factercheck, hardalov2022crowdchecked}.

\paragraph{Surveys on Automatic Fact-Checking.}
Recent work surveyed (aspects of) automated fact-checking and related tasks,
including 
explainability \citep{kotonya-toni-2020-explainable-survey}, 
stance classification \citep{kuecuek2020stance,hardalov2021survey}, 
propaganda detection \citep{da2021survey}, 
rumor detection on social media \citep{zubiaga2018detection, islam2020deep}, 
fake-news detection \citep{oshikawa-etal-2020-survey,zhou2020survey}, 
and automated fact-checking \citep{thorne-vlachos-2018-automated}. We refer interested readers to these papers.

\citet{guo2022survey} surveyed the state of automatic fact-checking.
Based on their work, we zoom in on real-world misinformation to investigate the gap between professional fact-checkers and \fcnlp{}.
Recently, \citet{nakov2021automated} surveyed tasks to assist humans during the verification. 
Our work differs in that we focus solely on the automatic verification approach of misinformation. They argue 
for the need for automatic tools to support \textit{humans} during verification.
Similarly, \citet{graves2018understanding} interviewed expert fact-checkers and computer scientists and conclude, that automatic fact-checking 
cannot replicate professional fact-checkers in the foreseeable future.
Our results confirm the challenging nature of misinformation but also outline why current models have unrealistic expectations, and how humans overcome these problems. We believe this to be important as real-world misinformation is well within the scope of current NLP research.

\paragraph{Towards Human Verification.}
A possible path forward is to align automatic verification with journalistic verification: 
Use the claimant's reasoning to find evidence and verify the claim.
This relies on the complex task of finding the correct sources \citep{arnold2020challenges}.
A fruitful but understudied direction may be automated provenance detection \citep{zhang-etal-2020-said,zhang-etal-2021-article}.
Building systems that can provide source guarantees paves the way for 
reasoning 
tasks, such as the detection of logical fallacies \citep{jin2022logical}, implicit warrants \citep{habernal-etal-2018-argument},
or propaganda techniques \citep{da-san-martino-etal-2019-fine,huang2022faking}.
Integrating sufficient context into datasets is non-trivial and may require tracing a claim and its source across multiple platforms. 
Existing literature shows the heterogeneity of misinformation \citep{borel2016chicago,wardle2017fake,cook2020deconstructing} and can help to identify small, focused problems that can realistically be translated into NLP.
Approaches from computer vision focus on misinformation-specific 
approaches to detect manipulated or misrepresented images \citep{zlatkova-etal-2019-fact,abdelnabi2021open,musi2022staying}. 

\section{Conclusion}
In this work, we contrasted NLP fact-checking approaches with how professional fact-checkers combat misinformation. We identified that reliance on counter-evidence hinders current fact-checking systems to refute real-world misinformation. 
Using \textsc{MultiFC} we find that most evidence is insufficient, or leaked and exploited by trained models.
Moving forward, we suggest to
align NLP approaches with the human verification process, and task definitions with smaller and well-defined verification strategies. 
\section*{Limitations}
The scope of this study is restricted to misinformation claims, and their representation as textual statements, that professional fact-checking organizations selected as important to verify. 
This only represents a fraction of all existing misinformation \citep{vinhas2022fact}.
Our findings cannot be generalized to other types of misinformation. Process definitions for claim selection and verification differ amongst fact-checkers \citep{arnold2020challenges}. 
The assessed claims for the analysis and experiments are biased to the claim selection criteria, including the domain, language, and geographical biases of Snopes and PolitiFact. 
Even fact-checkers cannot fully eliminate subjectivity. \citet{nieminen2021checking} find 11\% PolitiFact's verified claims uncheckable. We consider the fact-checkers assessment as the gold standard and adhere to the introduced subjectivity.
PolitiFact and Snopes verify claims from English-speaking countries with rich resources and trusted government documents. Fact-checking organizations may rely on different strategies, adapted to different scenarios such as different topics, dissemination of misinformation, or trust and availability of official information.\footnote{ \url{https://www.poynter.org/fact-checking/2019/heres-how-fact-checking-is-developing-across-africa/}}
The quantification of leaked evidence is bound to the time-frame, fact-checking organizations, and found evidence of \textsc{MultiFC}. We did not investigate the influence of different factors such as the fact-checkers language, domain, or popularity, nor did we evaluate different evidence collection strategies. The same restrictions apply to the experimental results. Further, following \citet{hansen2021automatic} we only consider labels on a veracity scale for the experiments (e.g. excluding ``misleading'').

\section*{Ethics Statement}
In this work we only consider publicly available data as provided by fact-checking organizations or \textsc{MultiFC}, but do not publish it. We do not use any personal data. 
We note that creating more realistic datasets (including realistic context), as suggested by us, induces ethical challenges as it requires personalized data (e.g. from Twitter or Facebook).
We consider this study's goal to reduce harmful misinformation by aligning automatic methods with best-practices from professional fact-checkers as ethically correct. 
However, even if successfully developed, fact-checking systems are inevitably imperfect. Malicious actors may design claims that exploit the system's weakness to predict the opposite verdict, giving legitimacy to false claims, or discrediting correct claims.
Further, malicious actors may develop fact-checking systems under their control. When extended with triggers enabling backdoor attacks \citep{chen2021badnl}  to control the outcome, these systems can serve as powerful tools to decide what seems true or false. 

\section*{Acknowledgements}
We thank the anonymous reviewers for their valuable feedback. We further thank Luke Bates, Tim Baumgärtner and Ilia Kuznetsov for their feedback on this work. 
 This research work has been funded by the German Federal Ministry of Education and Research and the Hessian Ministry of Higher Education, Research, Science and the Arts within their joint support of the National Research Center for Applied Cybersecurity ATHENE
 and by the European Regional Development Fund (ERDF) and the Hessian State Chancellery – Hessian Minister of Digital Strategy and Development under the promotional reference 20005482 (TexPrax).

\bibliography{main}

\begin{thebibliography}{83}
\expandafter\ifx\csname natexlab\endcsname\relax\def\natexlab#1{#1}\fi

\bibitem[{Abdelnabi et~al.(2022)Abdelnabi, Hasan, and
  Fritz}]{abdelnabi2021open}
Sahar Abdelnabi, Rakibul Hasan, and Mario Fritz. 2022.
\newblock \href
  {https://openaccess.thecvf.com/content/CVPR2022/html/Abdelnabi_Open-Domain_Content-Based_Multi-Modal_Fact-Checking_of_Out-of-Context_Images_via_Online_Resources_CVPR_2022_paper.html}
  {{Open-Domain, Content-based, Multi-modal Fact-checking of Out-of-Context
  Images via Online Resources}}.
\newblock In \emph{Proceedings of the IEEE/CVF Conference on Computer Vision
  and Pattern Recognition}, pages 14940--14949.

\bibitem[{Aghababaeian et~al.(2020)Aghababaeian, Hamdanieh, and
  Ostadtaghizadeh}]{aghababaeian2020alcohol}
Hamidreza Aghababaeian, Lara Hamdanieh, and Abbas Ostadtaghizadeh. 2020.
\newblock \href
  {https://www.sciencedirect.com/science/article/abs/pii/S0741832920302500}
  {{Alcohol intake in an attempt to fight COVID-19: A medical myth in Iran}}.
\newblock \emph{Alcohol}, 88:29--32.

\bibitem[{Alhindi et~al.(2018)Alhindi, Petridis, and
  Muresan}]{alhindi-etal-2018-evidence}
Tariq Alhindi, Savvas Petridis, and Smaranda Muresan. 2018.
\newblock \href {https://doi.org/10.18653/v1/W18-5513} {Where is your evidence:
  Improving fact-checking by justification modeling}.
\newblock In \emph{Proceedings of the First Workshop on Fact Extraction and
  {VER}ification ({FEVER})}, pages 85--90, Brussels, Belgium. Association for
  Computational Linguistics.

\bibitem[{Aly et~al.(2021)Aly, Guo, Schlichtkrull, Thorne, Vlachos,
  Christodoulopoulos, Cocarascu, and Mittal}]{aly2021feverous}
Rami Aly, Zhijiang Guo, Michael~Sejr Schlichtkrull, James Thorne, Andreas
  Vlachos, Christos Christodoulopoulos, Oana Cocarascu, and Arpit Mittal. 2021.
\newblock \href {https://openreview.net/forum?id=h-flVCIlstW} {{FEVEROUS: Fact
  Extraction and VERification Over Unstructured and Structured information}}.
\newblock In \emph{Thirty-fifth Conference on Neural Information Processing
  Systems Datasets and Benchmarks Track (Round 1)}.

\bibitem[{Arnold(2020)}]{arnold2020challenges}
Phoebe Arnold. 2020.
\newblock \href
  {https://fullfact.org/blog/2020/dec/the-challenges-of-online-fact-checking-how-technology-can-and-cant-help/}
  {The challenges of online fact checking: how technology can (and can’t)
  help}.
\newblock Technical report, FullFact.

\bibitem[{Augenstein et~al.(2019)Augenstein, Lioma, Wang, Chaves~Lima, Hansen,
  Hansen, and Simonsen}]{augenstein-etal-2019-multifc}
Isabelle Augenstein, Christina Lioma, Dongsheng Wang, Lucas Chaves~Lima, Casper
  Hansen, Christian Hansen, and Jakob~Grue Simonsen. 2019.
\newblock \href {https://doi.org/10.18653/v1/D19-1475} {{{M}ulti{FC}: A
  Real-World Multi-Domain Dataset for Evidence-Based Fact Checking of Claims}}.
\newblock In \emph{Proceedings of the 2019 Conference on Empirical Methods in
  Natural Language Processing and the 9th International Joint Conference on
  Natural Language Processing (EMNLP-IJCNLP)}, pages 4685--4697, Hong Kong,
  China. Association for Computational Linguistics.

\bibitem[{Baly et~al.(2018)Baly, Mohtarami, Glass, M{\`a}rquez, Moschitti, and
  Nakov}]{baly-etal-2018-integrating}
Ramy Baly, Mitra Mohtarami, James Glass, Llu{\'\i}s M{\`a}rquez, Alessandro
  Moschitti, and Preslav Nakov. 2018.
\newblock \href {https://doi.org/10.18653/v1/N18-2004} {{Integrating Stance
  Detection and Fact Checking in a Unified Corpus}}.
\newblock In \emph{Proceedings of the 2018 Conference of the North {A}merican
  Chapter of the Association for Computational Linguistics: Human Language
  Technologies, Volume 2 (Short Papers)}, pages 21--27, New Orleans, Louisiana.
  Association for Computational Linguistics.

\bibitem[{Borel(2016)}]{borel2016chicago}
Brooke Borel. 2016.
\newblock \emph{The Chicago guide to fact-checking}.
\newblock University of Chicago Press.

\bibitem[{Bovet and Makse(2019)}]{bovet2019influence}
Alexandre Bovet and Hern{\'a}n~A Makse. 2019.
\newblock \href {https://www.nature.com/articles/s41467-018-07761-2}
  {{Influence of fake news in Twitter during the 2016 US presidential
  election}}.
\newblock \emph{Nature communications}, 10(1):1--14.

\bibitem[{Bowman et~al.(2015)Bowman, Angeli, Potts, and
  Manning}]{bowman-etal-2015-large}
Samuel~R. Bowman, Gabor Angeli, Christopher Potts, and Christopher~D. Manning.
  2015.
\newblock \href {https://doi.org/10.18653/v1/D15-1075} {{A large annotated
  corpus for learning natural language inference}}.
\newblock In \emph{Proceedings of the 2015 Conference on Empirical Methods in
  Natural Language Processing}, pages 632--642, Lisbon, Portugal. Association
  for Computational Linguistics.

\bibitem[{Buttry(2014)}]{buttry2014verification}
Steve Buttry. 2014.
\newblock Verification fundamentals: Rules to live by.
\newblock \emph{Verification Handbook: A Definitive Guide to Verifying Digital
  Content for Emergency Coverage}, pages 15--23.

\bibitem[{Chen et~al.(2021)Chen, Salem, Chen, Backes, Ma, Shen, Wu, and
  Zhang}]{chen2021badnl}
Xiaoyi Chen, Ahmed Salem, Dingfan Chen, Michael Backes, Shiqing Ma, Qingni
  Shen, Zhonghai Wu, and Yang Zhang. 2021.
\newblock \href {https://dl.acm.org/doi/abs/10.1145/3485832.3485837} {{BadNL:
  Backdoor Attacks against NLP Models with Semantic-preserving Improvements}}.
\newblock In \emph{Annual Computer Security Applications Conference}, pages
  554--569.

\bibitem[{Cook(2020)}]{cook2020deconstructing}
John Cook. 2020.
\newblock \href
  {https://www.climatechangecommunication.org/wp-content/uploads/2020/11/Cook_2020_deconstructing_denial.pdf}
  {{Deconstructing Climate Science Denial}}.
\newblock In \emph{Edward Elgar Research Handbook in Communicating Climate
  Change}. Edward Elgar Publishing.

\bibitem[{Cui and Lee(2020)}]{cui2020coaid}
Limeng Cui and Dongwon Lee. 2020.
\newblock \href {https://arxiv.org/abs/2006.00885} {{CoAID: COVID-19 Healthcare
  Misinformation Dataset}}.
\newblock \emph{arXiv preprint arXiv:2006.00885}.

\bibitem[{Da~San~Martino et~al.(2021)Da~San~Martino, Cresci,
  Barr{\'o}n-Cede{\~n}o, Yu, Di~Pietro, and Nakov}]{da2021survey}
Giovanni Da~San~Martino, Stefano Cresci, Alberto Barr{\'o}n-Cede{\~n}o,
  Seunghak Yu, Roberto Di~Pietro, and Preslav Nakov. 2021.
\newblock \href {https://dl.acm.org/doi/10.5555/3491440.3492112} {A survey on
  computational propaganda detection}.
\newblock In \emph{Proceedings of the Twenty-Ninth International Conference on
  International Joint Conferences on Artificial Intelligence}, pages
  4826--4832.

\bibitem[{Da~San~Martino et~al.(2019)Da~San~Martino, Yu, Barr{\'o}n-Cede{\~n}o,
  Petrov, and Nakov}]{da-san-martino-etal-2019-fine}
Giovanni Da~San~Martino, Seunghak Yu, Alberto Barr{\'o}n-Cede{\~n}o, Rostislav
  Petrov, and Preslav Nakov. 2019.
\newblock \href {https://doi.org/10.18653/v1/D19-1565} {{Fine-Grained Analysis
  of Propaganda in News Article}}.
\newblock In \emph{Proceedings of the 2019 Conference on Empirical Methods in
  Natural Language Processing and the 9th International Joint Conference on
  Natural Language Processing (EMNLP-IJCNLP)}, pages 5636--5646, Hong Kong,
  China. Association for Computational Linguistics.

\bibitem[{Dadgar and Ghatee(2021)}]{dadgar2021checkovid}
Sajad Dadgar and Mehdi Ghatee. 2021.
\newblock \href {https://arxiv.org/abs/2107.09768} {{Checkovid: A COVID-19
  misinformation detection system on Twitter using network and content mining
  perspectives}}.
\newblock \emph{arXiv preprint arXiv:2107.09768}.

\bibitem[{Devlin et~al.(2019)Devlin, Chang, Lee, and
  Toutanova}]{devlin2018bert}
Jacob Devlin, Ming-Wei Chang, Kenton Lee, and Kristina Toutanova. 2019.
\newblock \href {https://doi.org/10.18653/v1/N19-1423} {{{BERT}: Pre-training
  of Deep Bidirectional Transformers for Language Understanding}}.
\newblock In \emph{Proceedings of the 2019 Conference of the North {A}merican
  Chapter of the Association for Computational Linguistics: Human Language
  Technologies, Volume 1 (Long and Short Papers)}, pages 4171--4186,
  Minneapolis, Minnesota. Association for Computational Linguistics.

\bibitem[{Diggelmann et~al.(2020)Diggelmann, Boyd-Graber, Bulian, Ciaramita,
  and Leippold}]{diggelmann2020climatefever}
Thomas Diggelmann, Jordan Boyd-Graber, Jannis Bulian, Massimiliano Ciaramita,
  and Markus Leippold. 2020.
\newblock {CLIMATE-FEVER: A Dataset for Verification of Real-World Climate
  Claims}.
\newblock In \emph{Tackling Climate Change with Machine Learning workshop at
  NeurIPS}.

\bibitem[{Eisenschlos et~al.(2021)Eisenschlos, Dhingra, Bulian,
  B{\"o}rschinger, and Boyd-Graber}]{eisenschlos-etal-2021-fool}
Julian Eisenschlos, Bhuwan Dhingra, Jannis Bulian, Benjamin B{\"o}rschinger,
  and Jordan Boyd-Graber. 2021.
\newblock \href {https://doi.org/10.18653/v1/2021.naacl-main.32} {Fool me
  twice: Entailment from {W}ikipedia gamification}.
\newblock In \emph{Proceedings of the 2021 Conference of the North American
  Chapter of the Association for Computational Linguistics: Human Language
  Technologies}, pages 352--365, Online. Association for Computational
  Linguistics.

\bibitem[{Ferreira and Vlachos(2016)}]{ferreira-vlachos-2016-emergent}
William Ferreira and Andreas Vlachos. 2016.
\newblock \href {https://doi.org/10.18653/v1/N16-1138} {{Emergent: a novel
  data-set for stance classification}}.
\newblock In \emph{Proceedings of the 2016 Conference of the North {A}merican
  Chapter of the Association for Computational Linguistics: Human Language
  Technologies}, pages 1163--1168, San Diego, California. Association for
  Computational Linguistics.

\bibitem[{Fisher et~al.(2016)Fisher, Cox, and Hermann}]{fisher2016pizzagate}
Marc Fisher, John~Woodrow Cox, and Peter Hermann. 2016.
\newblock {Pizzagate: From rumor, to hashtag, to gunfire in DC}.
\newblock \emph{Washington Post}, 6:8410--8415.

\bibitem[{FullFact(2020)}]{fullfact2020incident}
FullFact. 2020.
\newblock \href {https://fullfact.org/about/policy/incidentframework/report/}
  {Framework for information incidents}.
\newblock Technical report, FullFact.

\bibitem[{Golebiewski and Boyd(2019)}]{golebiewski2019data}
Michael Golebiewski and Danah Boyd. 2019.
\newblock \href {https://apo.org.au/node/265631} {{Data voids: Where missing
  data can easily be exploited}}.
\newblock Technical report, Data \& Society Research Institute.

\bibitem[{Graves(2018)}]{graves2018understanding}
Lucas Graves. 2018.
\newblock \href
  {https://reutersinstitute.politics.ox.ac.uk/our-research/understanding-promise-and-limits-automated-fact-checking}
  {{Understanding the Promise and Limits of Automated Fact-Checking}}.
\newblock In \emph{Reuters Institute for the Study of Journalism (Reuters
  Institute for the Study of Journalism Factsheets)}. Reuters Institute for the
  Study of Journalism.

\bibitem[{Guo et~al.(2022)Guo, Schlichtkrull, and Vlachos}]{guo2022survey}
Zhijiang Guo, Michael Schlichtkrull, and Andreas Vlachos. 2022.
\newblock \href {https://doi.org/10.1162/tacl_a_00454} {{A Survey on Automated
  Fact-Checking}}.
\newblock \emph{Transactions of the Association for Computational Linguistics},
  10:178--206.

\bibitem[{Gupta and Srikumar(2021)}]{gupta2021xfact}
Ashim Gupta and Vivek Srikumar. 2021.
\newblock \href {https://doi.org/10.18653/v1/2021.acl-short.86} {{{X}-Fact: A
  New Benchmark Dataset for Multilingual Fact Checking}}.
\newblock In \emph{Proceedings of the 59th Annual Meeting of the Association
  for Computational Linguistics and the 11th International Joint Conference on
  Natural Language Processing (Volume 2: Short Papers)}, pages 675--682,
  Online. Association for Computational Linguistics.

\bibitem[{Gupta et~al.(2022)Gupta, Wu, Liu, and Xiong}]{gupta2021dialfact}
Prakhar Gupta, Chien-Sheng Wu, Wenhao Liu, and Caiming Xiong. 2022.
\newblock \href {https://doi.org/10.18653/v1/2022.acl-long.263} {{{D}ial{F}act:
  A Benchmark for Fact-Checking in Dialogue}}.
\newblock In \emph{Proceedings of the 60th Annual Meeting of the Association
  for Computational Linguistics (Volume 1: Long Papers)}, pages 3785--3801,
  Dublin, Ireland. Association for Computational Linguistics.

\bibitem[{Habernal et~al.(2018)Habernal, Wachsmuth, Gurevych, and
  Stein}]{habernal-etal-2018-argument}
Ivan Habernal, Henning Wachsmuth, Iryna Gurevych, and Benno Stein. 2018.
\newblock \href {https://doi.org/10.18653/v1/N18-1175} {{The Argument Reasoning
  Comprehension Task: Identification and Reconstruction of Implicit Warrants}}.
\newblock In \emph{Proceedings of the 2018 Conference of the North {A}merican
  Chapter of the Association for Computational Linguistics: Human Language
  Technologies, Volume 1 (Long Papers)}, pages 1930--1940, New Orleans,
  Louisiana. Association for Computational Linguistics.

\bibitem[{Hanselowski et~al.(2019)Hanselowski, Stab, Schulz, Li, and
  Gurevych}]{hanselowski-etal-2019-richly}
Andreas Hanselowski, Christian Stab, Claudia Schulz, Zile Li, and Iryna
  Gurevych. 2019.
\newblock \href {https://doi.org/10.18653/v1/K19-1046} {{A Richly Annotated
  Corpus for Different Tasks in Automated Fact-Checking}}.
\newblock In \emph{Proceedings of the 23rd Conference on Computational Natural
  Language Learning (CoNLL)}, pages 493--503, Hong Kong, China. Association for
  Computational Linguistics.

\bibitem[{Hansen et~al.(2021)Hansen, Hansen, and
  Chaves~Lima}]{hansen2021automatic}
Casper Hansen, Christian Hansen, and Lucas Chaves~Lima. 2021.
\newblock \href {https://doi.org/10.18653/v1/2021.acl-short.12} {{Automatic
  Fake News Detection: Are Models Learning to Reason?}}
\newblock In \emph{Proceedings of the 59th Annual Meeting of the Association
  for Computational Linguistics and the 11th International Joint Conference on
  Natural Language Processing (Volume 2: Short Papers)}, pages 80--86, Online.
  Association for Computational Linguistics.

\bibitem[{Hardalov et~al.(2022{\natexlab{a}})Hardalov, Arora, Nakov, and
  Augenstein}]{hardalov2021survey}
Momchil Hardalov, Arnav Arora, Preslav Nakov, and Isabelle Augenstein.
  2022{\natexlab{a}}.
\newblock \href {https://doi.org/10.18653/v1/2022.findings-naacl.94} {{A Survey
  on Stance Detection for Mis- and Disinformation Identification}}.
\newblock In \emph{Findings of the Association for Computational Linguistics:
  NAACL 2022}, pages 1259--1277, Seattle, United States. Association for
  Computational Linguistics.

\bibitem[{Hardalov et~al.(2022{\natexlab{b}})Hardalov, Chernyavskiy, Koychev,
  Ilvovsky, and Nakov}]{hardalov2022crowdchecked}
Momchil Hardalov, Anton Chernyavskiy, Ivan Koychev, Dmitry Ilvovsky, and
  Preslav Nakov. 2022{\natexlab{b}}.
\newblock \href {https://arxiv.org/abs/2210.04447} {{CrowdChecked: Detecting
  Previously Fact-Checked Claims in Social Media}}.
\newblock In \emph{Proceedings of the 2nd Conference of the Asia-Pacific
  Chapter of the Association for Computational Linguistics and the 12th
  International Joint Conference on Natural Language Processing}, page (to
  appear), online. Association for Computational Linguistics.

\bibitem[{Hossain et~al.(2020)Hossain, Logan~IV, Ugarte, Matsubara, Young, and
  Singh}]{hossain-etal-2020-covidlies}
Tamanna Hossain, Robert~L. Logan~IV, Arjuna Ugarte, Yoshitomo Matsubara, Sean
  Young, and Sameer Singh. 2020.
\newblock \href {https://doi.org/10.18653/v1/2020.nlpcovid19-2.11}
  {{{COVIDL}ies: Detecting {COVID}-19 Misinformation on Social Media}}.
\newblock In \emph{Proceedings of the 1st Workshop on {NLP} for {COVID}-19
  (Part 2) at {EMNLP} 2020}, Online. Association for Computational Linguistics.

\bibitem[{Huang et~al.(2022)Huang, McKeown, Nakov, Choi, and
  Ji}]{huang2022faking}
Kung-Hsiang Huang, Kathleen McKeown, Preslav Nakov, Yejin Choi, and Heng Ji.
  2022.
\newblock \href {https://arxiv.org/abs/2203.05386} {{Faking Fake News for Real
  Fake News Detection: Propaganda-loaded Training Data Generation}}.
\newblock \emph{arXiv preprint arXiv:2203.05386}.

\bibitem[{Islam et~al.(2020)Islam, Liu, Wang, and Xu}]{islam2020deep}
Md~Rafiqul Islam, Shaowu Liu, Xianzhi Wang, and Guandong Xu. 2020.
\newblock \href {https://link.springer.com/article/10.1007/s13278-020-00696-x}
  {{Deep learning for misinformation detection on online social networks: a
  survey and new perspectives}}.
\newblock \emph{Social Network Analysis and Mining}, 10(1):1--20.

\bibitem[{Jiang et~al.(2020)Jiang, Bordia, Zhong, Dognin, Singh, and
  Bansal}]{jiang-etal-2020-hover}
Yichen Jiang, Shikha Bordia, Zheng Zhong, Charles Dognin, Maneesh Singh, and
  Mohit Bansal. 2020.
\newblock \href {https://doi.org/10.18653/v1/2020.findings-emnlp.309}
  {{{H}o{V}er: A Dataset for Many-Hop Fact Extraction And Claim Verification}}.
\newblock In \emph{Findings of the Association for Computational Linguistics:
  EMNLP 2020}, pages 3441--3460, Online. Association for Computational
  Linguistics.

\bibitem[{Jin et~al.(2022)Jin, Lalwani, Vaidhya, Shen, Ding, Lyu, Sachan,
  Mihalcea, and Sch{\"o}lkopf}]{jin2022logical}
Zhijing Jin, Abhinav Lalwani, Tejas Vaidhya, Xiaoyu Shen, Yiwen Ding, Zhiheng
  Lyu, Mrinmaya Sachan, Rada Mihalcea, and Bernhard Sch{\"o}lkopf. 2022.
\newblock \href {https://arxiv.org/abs/2202.13758} {{Logical Fallacy
  Detection}}.
\newblock \emph{arXiv preprint arXiv:2202.13758}.

\bibitem[{Khan et~al.(2022)Khan, Wang, and
  Poupart}]{khan-etal-2022-watclaimcheck}
Kashif Khan, Ruizhe Wang, and Pascal Poupart. 2022.
\newblock \href {https://aclanthology.org/2022.acl-long.92}
  {{W}at{C}laim{C}heck: A new dataset for claim entailment and inference}.
\newblock In \emph{Proceedings of the 60th Annual Meeting of the Association
  for Computational Linguistics (Volume 1: Long Papers)}, pages 1293--1304,
  Dublin, Ireland. Association for Computational Linguistics.

\bibitem[{Kotonya and
  Toni(2020{\natexlab{a}})}]{kotonya-toni-2020-explainable-survey}
Neema Kotonya and Francesca Toni. 2020{\natexlab{a}}.
\newblock \href {https://doi.org/10.18653/v1/2020.coling-main.474}
  {{Explainable Automated Fact-Checking: A Survey}}.
\newblock In \emph{Proceedings of the 28th International Conference on
  Computational Linguistics}, pages 5430--5443, Barcelona, Spain (Online).
  International Committee on Computational Linguistics.

\bibitem[{Kotonya and Toni(2020{\natexlab{b}})}]{kotonya-toni-2020-explainable}
Neema Kotonya and Francesca Toni. 2020{\natexlab{b}}.
\newblock \href {https://doi.org/10.18653/v1/2020.emnlp-main.623} {{Explainable
  Automated Fact-Checking for Public Health Claims}}.
\newblock In \emph{Proceedings of the 2020 Conference on Empirical Methods in
  Natural Language Processing (EMNLP)}, pages 7740--7754, Online. Association
  for Computational Linguistics.

\bibitem[{K\"{u}\c{c}\"{u}k and Can(2020)}]{kuecuek2020stance}
Dilek K\"{u}\c{c}\"{u}k and Fazli Can. 2020.
\newblock \href {https://doi.org/10.1145/3369026} {{Stance Detection: A
  Survey}}.
\newblock \emph{ACM Computing Surveys (CSUR)}, 53(1).

\bibitem[{Lewandowsky et~al.(2020)Lewandowsky, Cook, Ecker, Albarracin,
  Amazeen, Kendou, Lombardi, Newman, Pennycook, Porter, Rand, Rapp, Reifler,
  Roozenbeek, Schmid, Seifert, Sinatra, Swire-Thompson, van~der Linden, Vraga,
  Wood, and Zaragoza}]{lewandowsky2020debunking}
Stephan Lewandowsky, John Cook, Ullrich Ecker, Dolores Albarracin, Michelle
  Amazeen, P.~Kendou, D.~Lombardi, E.~Newman, G.~Pennycook, E.~Porter, D.~Rand,
  D.~Rapp, J.~Reifler, J.~Roozenbeek, P.~Schmid, C.~Seifert, G.~Sinatra,
  B.~Swire-Thompson, S.~van~der Linden, E.~Vraga, T.~Wood, and M.~Zaragoza.
  2020.
\newblock \href {https://open.bu.edu/handle/2144/43031} {\emph{{The Debunking
  Handbook 2020}}}.
\newblock OpenBU.

\bibitem[{Li et~al.(2020)Li, Jiang, Shu, and Liu}]{yichuan2020toward}
Yichuan Li, Bohan Jiang, Kai Shu, and Huan Liu. 2020.
\newblock \href {https://doi.org/10.1109/BigData50022.2020.9378472} {{Toward A
  Multilingual and Multimodal Data Repository for COVID-19 Disinformation}}.
\newblock In \emph{2020 IEEE International Conference on Big Data (Big Data)},
  pages 4325--4330.

\bibitem[{Mart{\'\i}n et~al.(2022)Mart{\'\i}n, Huertas-Tato,
  Huertas-Garc{\'\i}a, Villar-Rodr{\'\i}guez, and
  Camacho}]{martin2021factercheck}
Alejandro Mart{\'\i}n, Javier Huertas-Tato, {\'A}lvaro Huertas-Garc{\'\i}a,
  Guillermo Villar-Rodr{\'\i}guez, and David Camacho. 2022.
\newblock \href
  {https://www.sciencedirect.com/science/article/pii/S0950705122006323}
  {{FacTeR-Check: Semi-automated fact-checking through semantic similarity and
  natural language inference}}.
\newblock \emph{Knowledge-Based Systems}, 251:109265.

\bibitem[{Musi and Rocci(2022)}]{musi2022staying}
Elena Musi and Andrea Rocci. 2022.
\newblock \href {https://doi.org/10.1007/978-3-030-91017-4_16} {{Staying Up to
  Date with Fact and Reason Checking: An Argumentative Analysis of Outdated
  News}}.
\newblock In Steve Oswald, Marcin Lewi{\'{n}}ski, Sara Greco, and Serena
  Villata, editors, \emph{The Pandemic of Argumentation}, pages 311--330.
  Springer International Publishing, Cham.

\bibitem[{Nakov et~al.(2021)Nakov, Corney, Hasanain, Alam, Elsayed,
  Barrón-Cedeño, Papotti, Shaar, and Da~San~Martino}]{nakov2021automated}
Preslav Nakov, David Corney, Maram Hasanain, Firoj Alam, Tamer Elsayed, Alberto
  Barrón-Cedeño, Paolo Papotti, Shaden Shaar, and Giovanni Da~San~Martino.
  2021.
\newblock \href {https://doi.org/10.24963/ijcai.2021/619} {{Automated
  Fact-Checking for Assisting Human Fact-Checkers}}.
\newblock In \emph{Proceedings of the Thirtieth International Joint Conference
  on Artificial Intelligence, {IJCAI-21}}, pages 4551--4558. International
  Joint Conferences on Artificial Intelligence Organization.

\bibitem[{Nieminen and Sankari(2021)}]{nieminen2021checking}
Sakari Nieminen and Valtteri Sankari. 2021.
\newblock \href
  {https://www.tandfonline.com/doi/full/10.1080/1461670X.2021.1873818}
  {{Checking PolitiFact’s Fact-Checks}}.
\newblock \emph{Journalism Studies}, 22(3):358--378.

\bibitem[{Oshikawa et~al.(2020)Oshikawa, Qian, and
  Wang}]{oshikawa-etal-2020-survey}
Ray Oshikawa, Jing Qian, and William~Yang Wang. 2020.
\newblock \href {https://www.aclweb.org/anthology/2020.lrec-1.747} {{A Survey
  on Natural Language Processing for Fake News Detection}}.
\newblock In \emph{Proceedings of the 12th Language Resources and Evaluation
  Conference}, pages 6086--6093, Marseille, France. European Language Resources
  Association.

\bibitem[{Ostrowski et~al.(2021)Ostrowski, Arora, Atanasova, and
  Augenstein}]{ostrowski2020multi}
Wojciech Ostrowski, Arnav Arora, Pepa Atanasova, and Isabelle Augenstein. 2021.
\newblock \href {https://doi.org/10.24963/ijcai.2021/536} {{Multi-Hop Fact
  Checking of Political Claims}}.
\newblock In \emph{Proceedings of the Thirtieth International Joint Conference
  on Artificial Intelligence, {IJCAI-21}}, pages 3892--3898. International
  Joint Conferences on Artificial Intelligence Organization.

\bibitem[{Park et~al.(2022)Park, Min, Kang, Zettlemoyer, and
  Hajishirzi}]{park2021faviq}
Jungsoo Park, Sewon Min, Jaewoo Kang, Luke Zettlemoyer, and Hannaneh
  Hajishirzi. 2022.
\newblock \href {https://doi.org/10.18653/v1/2022.acl-long.354} {{{F}a{VIQ}:
  {FA}ct Verification from Information-seeking Questions}}.
\newblock In \emph{Proceedings of the 60th Annual Meeting of the Association
  for Computational Linguistics (Volume 1: Long Papers)}, pages 5154--5166,
  Dublin, Ireland. Association for Computational Linguistics.

\bibitem[{Patwa et~al.(2021)Patwa, Sharma, Pykl, Guptha, Kumari, Akhtar, Ekbal,
  Das, and Chakraborty}]{patwa2020fighting}
Parth Patwa, Shivam Sharma, Srinivas Pykl, Vineeth Guptha, Gitanjali Kumari,
  Md~Shad Akhtar, Asif Ekbal, Amitava Das, and Tanmoy Chakraborty. 2021.
\newblock \href {https://link.springer.com/chapter/10.1007/978-3-030-73696-5_3}
  {{Fighting an infodemic: Covid-19 fake news dataset}}.
\newblock In \emph{International Workshop on Combating Online Hostile Posts in
  Regional Languages during Emergency Situation}, pages 21--29. Springer.

\bibitem[{Pomerleau and Rao(2017)}]{pomerleau2018fake}
Dean Pomerleau and Delip Rao. 2017.
\newblock {Fake News Challenge}.
\newblock \url{http://www.fakenewschallenge.org/}.

\bibitem[{Rashkin et~al.(2017)Rashkin, Choi, Jang, Volkova, and
  Choi}]{rashkin-etal-2017-truth}
Hannah Rashkin, Eunsol Choi, Jin~Yea Jang, Svitlana Volkova, and Yejin Choi.
  2017.
\newblock \href {https://doi.org/10.18653/v1/D17-1317} {{Truth of Varying
  Shades: Analyzing Language in Fake News and Political Fact-Checking}}.
\newblock In \emph{Proceedings of the 2017 Conference on Empirical Methods in
  Natural Language Processing}, pages 2931--2937, Copenhagen, Denmark.
  Association for Computational Linguistics.

\bibitem[{Saakyan et~al.(2021)Saakyan, Chakrabarty, and
  Muresan}]{saakyan2021covidfact}
Arkadiy Saakyan, Tuhin Chakrabarty, and Smaranda Muresan. 2021.
\newblock \href {https://doi.org/10.18653/v1/2021.acl-long.165} {{{COVID}-Fact:
  Fact Extraction and Verification of Real-World Claims on {COVID}-19
  Pandemic}}.
\newblock In \emph{Proceedings of the 59th Annual Meeting of the Association
  for Computational Linguistics and the 11th International Joint Conference on
  Natural Language Processing (Volume 1: Long Papers)}, pages 2116--2129,
  Online. Association for Computational Linguistics.

\bibitem[{Sarrouti et~al.(2021)Sarrouti, Ben~Abacha, Mrabet, and
  Demner-Fushman}]{sarrouti-etal-2021-evidence-based}
Mourad Sarrouti, Asma Ben~Abacha, Yassine Mrabet, and Dina Demner-Fushman.
  2021.
\newblock \href {https://doi.org/10.18653/v1/2021.findings-emnlp.297}
  {{Evidence-based Fact-Checking of Health-related Claims}}.
\newblock In \emph{Findings of the Association for Computational Linguistics:
  EMNLP 2021}, pages 3499--3512, Punta Cana, Dominican Republic. Association
  for Computational Linguistics.

\bibitem[{Sathe et~al.(2020)Sathe, Ather, Le, Perry, and
  Park}]{sathe-etal-2020-automated}
Aalok Sathe, Salar Ather, Tuan~Manh Le, Nathan Perry, and Joonsuk Park. 2020.
\newblock \href {https://www.aclweb.org/anthology/2020.lrec-1.849} {Automated
  fact-checking of claims from {W}ikipedia}.
\newblock In \emph{Proceedings of the 12th Language Resources and Evaluation
  Conference}, pages 6874--6882, Marseille, France. European Language Resources
  Association.

\bibitem[{Schuster et~al.(2019)Schuster, Shah, Yeo, Roberto Filizzola~Ortiz,
  Santus, and Barzilay}]{schuster-etal-2019-towards}
Tal Schuster, Darsh Shah, Yun Jie~Serene Yeo, Daniel Roberto Filizzola~Ortiz,
  Enrico Santus, and Regina Barzilay. 2019.
\newblock \href {https://doi.org/10.18653/v1/D19-1341} {{Towards Debiasing Fact
  Verification Models}}.
\newblock In \emph{Proceedings of the 2019 Conference on Empirical Methods in
  Natural Language Processing and the 9th International Joint Conference on
  Natural Language Processing (EMNLP-IJCNLP)}, pages 3419--3425, Hong Kong,
  China. Association for Computational Linguistics.

\bibitem[{Shaar et~al.(2020)Shaar, Babulkov, Da~San~Martino, and
  Nakov}]{shaar-etal-2020-known}
Shaden Shaar, Nikolay Babulkov, Giovanni Da~San~Martino, and Preslav Nakov.
  2020.
\newblock \href {https://doi.org/10.18653/v1/2020.acl-main.332} {{That is a
  Known Lie: Detecting Previously Fact-Checked Claims}}.
\newblock In \emph{Proceedings of the 58th Annual Meeting of the Association
  for Computational Linguistics}, pages 3607--3618, Online. Association for
  Computational Linguistics.

\bibitem[{Shane and Noel(2020)}]{shane2020data}
Tommy Shane and Pedro Noel. 2020.
\newblock \href {https://firstdraftnews.org/long-form-article/data-deficits/}
  {Data deficits: why we need to monitor the demand and supply of information
  in real time}.
\newblock Technical report, First Draft.

\bibitem[{Silverman(2014)}]{silverman2014verification}
Craig Silverman. 2014.
\newblock \href
  {https://verificationhandbook.com/downloads/verification.handbook.pdf}
  {\emph{Verification handbook: An ultimate guideline on digital age sourcing
  for emergency coverage}}.
\newblock European Journalism Centre.

\bibitem[{Silverman(2016)}]{silverman2016verification}
Craig Silverman. 2016.
\newblock \href
  {https://datajournalism.com/read/handbook/verification-1/additional-materials/verification-and-fact-checking}
  {\emph{Verification handbook: Additional Materials}}.
\newblock European Journalism Centre.

\bibitem[{Simon et~al.(2020)Simon, Howard, and Nielson}]{brennen2020types}
Felix Simon, Philip~N. Howard, and Rasmus~Kleis Nielson. 2020.
\newblock \href
  {https://reutersinstitute.politics.ox.ac.uk/types-sources-and-claims-covid-19-misinformation}
  {{Types, sources, and claims of COVID-19 misinformation}}.
\newblock Technical report, Reuters Institute for the Study of Journalism.

\bibitem[{Thorne et~al.(2021)Thorne, Glockner, Vallejo, Vlachos, and
  Gurevych}]{thorne2021evidence}
James Thorne, Max Glockner, Gisela Vallejo, Andreas Vlachos, and Iryna
  Gurevych. 2021.
\newblock \href {https://arxiv.org/abs/2104.00640} {{Evidence-based
  Verification for Real World Information Needs}}.
\newblock \emph{arXiv preprint arXiv:2104.00640}.

\bibitem[{Thorne and Vlachos(2018)}]{thorne-vlachos-2018-automated}
James Thorne and Andreas Vlachos. 2018.
\newblock \href {https://www.aclweb.org/anthology/C18-1283} {{Automated Fact
  Checking: Task Formulations, Methods and Future Directions}}.
\newblock In \emph{Proceedings of the 27th International Conference on
  Computational Linguistics}, pages 3346--3359, Santa Fe, New Mexico, USA.
  Association for Computational Linguistics.

\bibitem[{Thorne et~al.(2018)Thorne, Vlachos, Christodoulopoulos, and
  Mittal}]{thorne-etal-2018-fever}
James Thorne, Andreas Vlachos, Christos Christodoulopoulos, and Arpit Mittal.
  2018.
\newblock \href {https://doi.org/10.18653/v1/N18-1074} {{FEVER: a Large-scale
  Dataset for Fact Extraction and {VER}ification}}.
\newblock In \emph{Proceedings of the 2018 Conference of the North {A}merican
  Chapter of the Association for Computational Linguistics: Human Language
  Technologies, Volume 1 (Long Papers)}, pages 809--819, New Orleans,
  Louisiana. Association for Computational Linguistics.

\bibitem[{Urbani(2020)}]{urbani2020verifying}
Shaydanay Urbani. 2020.
\newblock \href
  {https://firstdraftnews.org/long-form-article/verifying-online-information/}
  {{Verifying Online Information}}.
\newblock Technical report, First Draft.

\bibitem[{van~der Linden(2022)}]{van2022misinformation}
Sander van~der Linden. 2022.
\newblock \href {https://www.nature.com/articles/s41591-022-01713-6}
  {Misinformation: susceptibility, spread, and interventions to immunize the
  public}.
\newblock \emph{Nature Medicine}, 28(3):460--467.

\bibitem[{Vinhas and Bastos(2022)}]{vinhas2022fact}
Ot{\'a}vio Vinhas and Marco Bastos. 2022.
\newblock \href
  {https://www.tandfonline.com/doi/abs/10.1080/1461670X.2022.2031259}
  {{Fact-Checking Misinformation: Eight Notes on Consensus Reality}}.
\newblock \emph{Journalism Studies}, 23(4):448--468.

\bibitem[{Vlachos and Riedel(2014)}]{vlachos-riedel-2014-fact}
Andreas Vlachos and Sebastian Riedel. 2014.
\newblock \href {https://doi.org/10.3115/v1/W14-2508} {{Fact Checking: Task
  definition and dataset construction}}.
\newblock In \emph{Proceedings of the {ACL} 2014 Workshop on Language
  Technologies and Computational Social Science}, pages 18--22, Baltimore, MD,
  USA. Association for Computational Linguistics.

\bibitem[{Vo and Lee(2020)}]{vo-lee-2020-facts}
Nguyen Vo and Kyumin Lee. 2020.
\newblock \href {https://doi.org/10.18653/v1/2020.emnlp-main.621} {{Where Are
  the Facts? Searching for Fact-checked Information to Alleviate the Spread of
  Fake News}}.
\newblock In \emph{Proceedings of the 2020 Conference on Empirical Methods in
  Natural Language Processing (EMNLP)}, pages 7717--7731, Online. Association
  for Computational Linguistics.

\bibitem[{Wadden et~al.(2020)Wadden, Lin, Lo, Wang, van Zuylen, Cohan, and
  Hajishirzi}]{wadden-etal-2020-fact}
David Wadden, Shanchuan Lin, Kyle Lo, Lucy~Lu Wang, Madeleine van Zuylen, Arman
  Cohan, and Hannaneh Hajishirzi. 2020.
\newblock \href {https://doi.org/10.18653/v1/2020.emnlp-main.609} {{Fact or
  Fiction: Verifying Scientific Claims}}.
\newblock In \emph{Proceedings of the 2020 Conference on Empirical Methods in
  Natural Language Processing (EMNLP)}, pages 7534--7550, Online. Association
  for Computational Linguistics.

\bibitem[{Wang(2017)}]{wang-2017-liar}
William~Yang Wang. 2017.
\newblock \href {https://doi.org/10.18653/v1/P17-2067} {{``Liar, Liar Pants on
  Fire'': A New Benchmark Dataset for Fake News Detection}}.
\newblock In \emph{Proceedings of the 55th Annual Meeting of the Association
  for Computational Linguistics (Volume 2: Short Papers)}, pages 422--426,
  Vancouver, Canada. Association for Computational Linguistics.

\bibitem[{Wardle et~al.(2017)}]{wardle2017fake}
Claire Wardle et~al. 2017.
\newblock \href {https://firstdraftnews.org/articles/fake-news-complicated/}
  {Fake news. it’s complicated}.
\newblock Technical report, First Draft.

\bibitem[{Weinzierl and Harabagiu(2022)}]{weinzierl2022vaccinelies}
Maxwell Weinzierl and Sanda Harabagiu. 2022.
\newblock \href {https://aclanthology.org/2022.lrec-1.753} {{V}accine{L}ies: A
  natural language resource for learning to recognize misinformation about the
  {COVID}-19 and {HPV} vaccines}.
\newblock In \emph{Proceedings of the Thirteenth Language Resources and
  Evaluation Conference}, pages 6967--6975, Marseille, France. European
  Language Resources Association.

\bibitem[{Wolf et~al.(2020)Wolf, Debut, Sanh, Chaumond, Delangue, Moi, Cistac,
  Rault, Louf, Funtowicz, Davison, Shleifer, von Platen, Ma, Jernite, Plu, Xu,
  Le~Scao, Gugger, Drame, Lhoest, and Rush}]{wolf-etal-2020-transformers}
Thomas Wolf, Lysandre Debut, Victor Sanh, Julien Chaumond, Clement Delangue,
  Anthony Moi, Pierric Cistac, Tim Rault, Remi Louf, Morgan Funtowicz, Joe
  Davison, Sam Shleifer, Patrick von Platen, Clara Ma, Yacine Jernite, Julien
  Plu, Canwen Xu, Teven Le~Scao, Sylvain Gugger, Mariama Drame, Quentin Lhoest,
  and Alexander Rush. 2020.
\newblock \href {https://doi.org/10.18653/v1/2020.emnlp-demos.6}
  {{Transformers: State-of-the-Art Natural Language Processing}}.
\newblock In \emph{Proceedings of the 2020 Conference on Empirical Methods in
  Natural Language Processing: System Demonstrations}, pages 38--45, Online.
  Association for Computational Linguistics.

\bibitem[{Zarocostas(2020)}]{zarocostas2020fight}
John Zarocostas. 2020.
\newblock \href {https://doi.org/https://doi.org/10.1016/S0140-6736(20)30461-X}
  {{How to fight an infodemic}}.
\newblock \emph{The Lancet}, 395(10225):676.

\bibitem[{Zhang et~al.(2020)Zhang, Ives, and Roth}]{zhang-etal-2020-said}
Yi~Zhang, Zachary Ives, and Dan Roth. 2020.
\newblock \href {https://doi.org/10.18653/v1/2020.acl-main.406} {{{``}Who said
  it, and Why?{''} Provenance for Natural Language Claims}}.
\newblock In \emph{Proceedings of the 58th Annual Meeting of the Association
  for Computational Linguistics}, pages 4416--4426, Online. Association for
  Computational Linguistics.

\bibitem[{Zhang et~al.(2021)Zhang, Ives, and Roth}]{zhang-etal-2021-article}
Yi~Zhang, Zachary Ives, and Dan Roth. 2021.
\newblock \href {https://doi.org/10.18653/v1/2021.acl-long.458} {{What is Your
  Article Based On? Inferring Fine-grained Provenance}}.
\newblock In \emph{Proceedings of the 59th Annual Meeting of the Association
  for Computational Linguistics and the 11th International Joint Conference on
  Natural Language Processing (Volume 1: Long Papers)}, pages 5894--5903,
  Online. Association for Computational Linguistics.

\bibitem[{Zhou and Zafarani(2020)}]{zhou2020survey}
Xinyi Zhou and Reza Zafarani. 2020.
\newblock \href {https://dl.acm.org/doi/abs/10.1145/3395046} {{A survey of fake
  news: Fundamental theories, detection methods, and opportunities}}.
\newblock \emph{ACM Computing Surveys (CSUR)}, 53(5):1--40.

\bibitem[{Zlatkova et~al.(2019)Zlatkova, Nakov, and
  Koychev}]{zlatkova-etal-2019-fact}
Dimitrina Zlatkova, Preslav Nakov, and Ivan Koychev. 2019.
\newblock \href {https://doi.org/10.18653/v1/D19-1216} {{Fact-Checking Meets
  Fauxtography: Verifying Claims About Images}}.
\newblock In \emph{Proceedings of the 2019 Conference on Empirical Methods in
  Natural Language Processing and the 9th International Joint Conference on
  Natural Language Processing (EMNLP-IJCNLP)}, pages 2099--2108, Hong Kong,
  China. Association for Computational Linguistics.

\bibitem[{Zubiaga et~al.(2018)Zubiaga, Aker, Bontcheva, Liakata, and
  Procter}]{zubiaga2018detection}
Arkaitz Zubiaga, Ahmet Aker, Kalina Bontcheva, Maria Liakata, and Rob Procter.
  2018.
\newblock \href {https://doi.org/10.1145/3161603} {{Detection and Resolution of
  Rumours in Social Media: A Survey}}.
\newblock \emph{ACM Computing Surveys (CSUR)}, 51(2).

\bibitem[{Zubiaga et~al.(2016)Zubiaga, Kochkina, Liakata, Procter, and
  Lukasik}]{zubiaga-etal-2016-stance}
Arkaitz Zubiaga, Elena Kochkina, Maria Liakata, Rob Procter, and Michal
  Lukasik. 2016.
\newblock \href {https://www.aclweb.org/anthology/C16-1230} {{Stance
  Classification in Rumours as a Sequential Task Exploiting the Tree Structure
  of Social Media Conversations}}.
\newblock In \emph{Proceedings of {COLING} 2016, the 26th International
  Conference on Computational Linguistics: Technical Papers}, pages 2438--2448,
  Osaka, Japan. The COLING 2016 Organizing Committee.

\end{thebibliography}
\bibliographystyle{acl_natbib}

\appendix
\section{Human Misinformation Verification Examples}
\label{appendix:human-misinformation:examples}

\begin{table*}[]
\small
    \centering
    \begin{tabularx}{\textwidth}{c c X c }
    \toprule
 \textbf{\#} & \textbf{Year} & \textbf{Misinformation Claim} & \textbf{Strategy} \\
    \toprule
    
(1) & 2020 & Tennis star Serena Williams posted a message on social media that began, ``I'm sick of COVID-19. I'm sick of black vs. white. I'm sick of Republicans vs. Democrats.'' & \textit{n/a} \\
    
(2) & 2010 & You can look up anyone's driver's license for free through the 'National Motor Vehicle Licence Organization' web site. & \textit{n/a} \\

(3) & 2016 & A photograph shows a newly hatched baby dragon. & \textit{n/a} \\
    
(4) & 2018 & Couple Arrested For Selling ‘Golden Tickets To Heaven. & \textit{n/a} \\

    \midrule
    
(5) & 2021 & There is no added safety to the public if you're vaccinated. & \textit{GCE}  \\

(6) &  2011 & Limiting labor negotiations to only wages is how it is for the most part in the private sector. &  \textit{GCE}  \\

(7) & 2018 & Kathy Manning gave nearly \$1 million to liberals & \textit{GCE}  \\

(8) & 2018 & Democrats let him (cop killer Luis Bracamontes) into our country, and Democrats let him stay. & \textit{GCE}  \\

    \midrule
(9) & 2014 &  A list reproduces Saul Alinsky's rules for ``How to Create a Social State.". & \textit{LCE}  \\
(10) & 2014 & Greg Abbott and his surrogates have referred to women who have been the victims of rape or incest as though somehow what they are confronting is a minor issue. & \textit{LCE}  \\

(11) & 2021 & During protests over the police in-custody death of George Floyd in the summer of 2020, Kamala Harris donated money to a Minnesota nonprofit that helped protesters who were arrested get out of jail and break more laws. & \textit{LCE}\\
    
    \bottomrule
     
    \end{tabularx}
    \caption{Example claims from Snopes and PolitiFact that professional fact-checkers refuted with global counter-evidence (\textit{GCE)}, local counter-evidence (\textit{LCE}) via the source guarantee, or marked as not-applicable (\textit{n/a)} for \fcnlp{}.}
    \label{tab:example-claims-theoretical-analysis-appendix}
\end{table*}
Table~\ref{tab:example-claims-theoretical-analysis-appendix} shows further examples for the reasoning over global counter-evidence, or when source guarantees are required. Further, we list examples we considered inapplicable to \fcnlp{} by nature and ergo excluded. Cases include imposter content (1), fake web pages (2), and claims that are about multi-modal content (3) or require reasoning over multi-modal sources during verification (4).
We show further examples of complex and different reasoning via global counter-evidence. This includes:
reasoning over scientific documents (5);
aggregating counter-examples (6); 
aggregating  distinct donations (7);
finding and contextualizing multiple events, including deportation and imprisonment of a murderer, and aligning these events with the political leadership in the U.S. (8);
Claims (9--11) require source guarantees to resources like a specific document (9), event (10), or organization (11).
\section{Leaked Evidence Analysis}
\subsection{Misinformation Labels}
\label{app:multifc-analysis:misinformation}
We consider all claims rated with strongly-leaning false verdicts and other verdicts that fall into the misinformation category such as ``misleading'' as misinformation. Remaining claims are either true (e.g. ``verified'', ``mostly true''), mixed (e.g. ``half-true'', ``outdated'') or not clearly applicable to misinformation (e.g. ``opinion!'', ``scam'', ``full flop'').
We provide all considered labels within \textsc{MultiFC} below.
\begin{itemize}
\small
    \item \textit{ABC}: in-the-red
    \item \textit{Africa Check}: incorrect, misleading
    \item \textit{BOOM Live}: rating: false
    \item \textit{Check Your Fact}: verdict: false
    \item \textit{Climate Feedback}: incrorrect, misleading
    \item \textit{FactScan}: factscan score: false, factscan score: misleading
    \item \textit{Factly}: false
    \item \textit{FactCheckNI}: conclusion: false
    \item \textit{FactCheck.org}: false, distorts the facts, misleading, spins the facts, not the whole story, cherry picks
    \item \textit{Gossip Cop}: 0, 1, 2, 3
    \item \textit{Hoax Slayer}: fake news
    \item \textit{Huffington Post CA}: a lot of baloney
    \item \textit{MPR News}: false, misleading
    \item \textit{Observatory}: mostly\_false
    \item \textit{Pandora}: mostly false, false, pants on fire!
    \item \textit{PesaCheck}: false
    \item \textit{PolitiFact}: mostly false,  false, pants on fire!, fiction
    \item \textit{Radio NZ}: fiction
    \item \textit{Snopes}: false, mostly false, miscaptioned, misattributed
    \item \textit{The Ferret}: mostly false, false
    \item \textit{The Journal}: we rate this claim false
    \item \textit{Truth Or Fiction}: fiction!, mostly fiction!, incorrect attribution!, misleading!, inaccurate attribution!
    \item \textit{VERA Files}: fake, misleading, false
    \item \textit{Voice of San Diego}: determination: false, determination: huckster propaganda, determination: barely true, determination: misleading
    \item \textit{Washington Post}: 4 pinnochios, false, not the whole story, needs context
\end{itemize}

\subsection{Automatic Identification of Leaked Evidence}
\label{app:multifc-analysis:misinformation:automatic-leak-detection}
\begin{table*}[]
\small
    \centering
    \begin{tabular}{l|l}
     \textbf{Organization}    &  \textbf{Template}\\
     \toprule
      Africa Check   & \texttt{africacheck.org/reports} \\
      AFP Fact Check   & \texttt{factcheck.afp.com} \\
        Check Your Fact   & \texttt{checkyourfact.com} \\
      Climate Feedback   & \texttt{climatefeedback.org/claimreview} \\
      Fact or Fiction   & \texttt{radionz.co.nz/programmes/election17-fact-or-fiction} \\
      FactCheck.org   & \texttt{factcheck.org} \\
      FactCheckNI   & \texttt{factcheckni.org} \\
      FACTLY   & \texttt{factly.in} \\
      FactsCan   & \texttt{factscan.ca} \\
      Full Fact   & \texttt{fullfact.org} \\
      Gossip Cop   & \texttt{gossipcop.com} \\
      Health Feedback   & \texttt{healthfeedback.org/claimreview} \\
    Hoax Slayer   & \texttt{hoax-slayer.net} \\
      Lead Stories FactChecker & \texttt{hoax-alert.leadstories.com} \\
      PesaCheck   & \texttt{pesacheck.org} \\
      PolitiFact   & \texttt{politifact.com} \\
      Snopes   & \texttt{snopes.com} \\
      Truth or Fiction   & \texttt{truthorfiction.com} \\
      Washington Post  & \texttt{washingtonpost.com/news/fact-checker} \\
    \end{tabular}
    \caption{Used Templates to automatically identify leaked snippets via the URLs.}
    \label{tab:automatically-identified-leaked-evidence:urls}
\end{table*}
\begin{table*}[]
\small
    \centering
    \begin{tabularx}{\textwidth}{l|X}
        \textbf{Regular Expression} & \textbf{Example} \\
        \toprule
        \texttt{'\textasciicircum false:'} &  \highlight{FALSE:} Map Shows Results of the 2012 Presidential Election If Only ... A map doesn't show the results of the 2012 election if only people who pay ...  FALSE: Map Shows Results of the 2012 Presidential Election If Only Taxpayers  Had ... is a map of how the Electoral College vote would look like if ONLY those  who ... \\
        
        \texttt{'politifact'} & \highlight{PolitiFact}: Testing Kathleen Vinehout claim on Scott Walker, new car ... Dec 20, 2013 ... We check a claim by state Sen. Kathleen Vinehout that Gov. Scott Walker bought  "80 new, brand new vehicles" that "we probably don't need.".\\
        
        \texttt{'snopes'} &  Real History Blog: The ACLU has NOT filed suit to have all military ... Feb 10, 2010 ... The ACLU has never filed such a suit, says the ACLU. Says \highlight{Snopes}, if ... and  another suit to end prayer from the military completely. They're ... \\
        
        \texttt{'\textasciicircum debunk'} & \highlight{Debunk}ed: Did 'The Simpsons' predict President Donald Trump's ... Feb 9, 2017 ... 'The Simpsons' has predicted a number of world events and an internet rumor  said the show predicted the death of Donald Trump. Veuer's Nick ... \\
        
        \texttt{'real story behind'} & \highlight{The real story} behind the statistic Trump just used to attack Obamacare; Jun 13, 2017 ... ... tweeted that 2 million people “just dropped out of ObamaCare.” 2 million more  people just dropped out of ObamaCare. It is in a death spiral.\\
        
        \texttt{'\textbackslash bfake\textbackslash b'} & Trump "moron" Harley-Davidson CEO quote: \highlight{Fake}.Jun 27, 2018 ... The CEO of Harley-Davidson Did Not Call Donald Trump a “Moron” ... Harley  Davidson CEO Matthew S Levatich says: "Our decision to move ...\\
        
        \texttt{'\textbackslash bhoax\textbackslash b'} & Eddie Murphy - latest news, breaking stories and comment - The ... All the latest breaking news on Eddie Murphy. Browse The ... Paul Walker  tragedy sparks Eddie Murphy Twitter death \highlight{hoax} · News · Final film of the Twilight  ...\\
        
        \texttt{'\textbackslash bfalsely\textbackslash b'} & CNN helpfully fact-checks Donald Trump's tweet about its “way down”; Jun 27, 2017 ... Trump tweeted that “Fake News CNN” had its “Ratings way down!” which he said  was due to the network being “caught \highlight{falsely} pushing their ...\\
        
        \texttt{'\textbackslash brumors?\textbackslash b'} & Did the Obama White House ban Christmas Nativity scenes ... Nov 21, 2018 ... Contrary to "War on Christmas" \highlight{rumors}, the Obama White House did not ban  Nativity scenes from the premises: ...\\
        
        \texttt{'\textbackslash bmyths?\textbackslash b'} & Did Coca-Cola Contain Coke? Here's What History Says; Since I was a little girl, I've heard the \highlight{myth} that Coca-Cola used to actually  contain cocaine. However, how credible is this rumor? I set out to find if there was  any ...\\
        
        \texttt{'\textbackslash bnot real news\textbackslash b'} & \highlight{NOT REAL NEWS}: A look at what didn't happen this week; Aug 11, 2017 ... NOT REAL: John McCain Says He 'Accidentally' Voted No On Healthcare Repeal  ... last month that sank a GOP effort to repeal the Affordable Care Act, ... story  purportedly showing the fake senator in handcuffs is actually a ...\\
        \texttt{'\textbackslash bunfounded\textbackslash b'} & No, 15,000 people did not vote for Harambe in 2016 | PunditFact; Nov 22, 2016 ... Harambe received 15,000 votes in the presidential election. ... Rumors that  15,000 people voted for the dead gorilla Harambe are \highlight{unfounded}.\\
        \texttt{'fact[ -]check'} & \highlight{Fact-check}ing an immigration meme that's been circulating for more ...Jul 5, 2018 ... "More than 66\% of ALL births in California are to illegals on Medi-Cal" ...  According to Medi-Cal, 50.4 percent of the state's births that year were ...\\
    \end{tabularx}
    \caption{Used regular expressions to automatically identify leaked evidence snippets.}
    \label{tab:automatically-identified-leaked-evidence:phrases}
\end{table*}
Table~\ref{tab:automatically-identified-leaked-evidence:urls} shows the URLs used to automatically determine leaked evidence. We consider an evidence snippet as leaked if any URLs of Table~\ref{tab:automatically-identified-leaked-evidence:urls} is a substring of the snippet's URL. We exclude URLs if they may also cover URLs to news articles.
Further, we consider an evidence snippet as leaked, if its lowercased title or text matches any of the regular expressions in Table~\ref{tab:automatically-identified-leaked-evidence:phrases}.
We identified two commonly made errors:
\begin{itemize}
    \item \textbf{Different Claim:} The approach considered evidence as leaked if it is not relevant to the exact same claim, but connected to the 
    same incident (\textit{``President Obama pushed through the stimulus based on the promise of keeping unemployment under 8 percent.''} and \textit{``The president promised that if he spent money on a stimulus program that unemployment would go to 5.7 percent or 6 percent. Those were his words''}), 
    or thematically related (\textit{``Cadbury chocolate eggs are infected with HIV-positive blood''} and \textit{``HIV \& AIDS infected oranges coming from Libya''}).
    \item \textbf{Discussing Fake News or Fact-Checking:} The approach selects snippets, that discuss fact-checking or fake news from a different perspective, not as a result of verification. This includes opinions or reports complaining about ``fake news'' being spread, or about the fact-checking process.
\end{itemize}
\subsection{Manual Guidelines}
\label{appendix:multifc-analysis:leaked-guidelines}
To determine the stance of evidence snippets to a claim, or whether it is leaked or not, we proceed in the following order: 
We first read the original fact-checking article, to fully comprehend the claim and how fact-checkers refuted it. 
If the title or text of the evidence snippets provides sufficient information, we decide based on the snippet alone.
If we cannot make a clear decision based on the snippet, we consider the original web page. This may be required, as evidence snippets often contain incomplete sentences.

\subsubsection{Leaked Evidence Snippets}
We consider evidence snippets as leaked if 
(a) they constitute information that relies on the verification of the same claim, 
or (b) provides originally unknown information from the claim's future.
When relying on content from the claim's verification, we do not require the information to contradict the claim from a human perspective. This often occurs when different pages (such as overview pages) reference the fact-checking article. Such a page may be a clear indication of the verdict in some cases (e.g. if titled ``All False Claims by Person A''). In other cases, different interpretations are valid: The statement ``We previously fact-checked similar claims that ...'' may be seen as neutral or as a give-away that similar claims were refuted. Further, humans cannot judge whether models may rely on latent patterns. An overview page titled ``All claims from Person A'' may be sufficient for the model if it learned that most claims by Person A are false. To remove this ambiguity, we consider any mention / or information taken form the claim's verification as leaked.
We do not strictly consider all evidence that appeared after the verification as leaked. Not  all evidence published after the claim's verification, is based not based on the verification. If not, we verify whether it relies on new information that previously did not exist or whether the truth changed. Consider the claim ``Khloe Kardashian did give birth over easter.'' refuted on April 5, 2018. Evidence about her actual birth on April 12, 2018, does not rely on a previous verification but is still considered leaked (new information available). In other cases (``Coca-Cola's "Share A Coke" campaign includes a bottle for the KKK.'', March 2, 2016) we consider evidence from March 30, 2016, as unleaked: It correctly reports about the same incident the claim refers to without any mention of the false claim, or its verification.

\subsubsection{Stance of Evidence Snippets}
For most claims, it is unrealistic to assume a single evidence snippet can refute them entirely. We follow \citet{sarrouti-etal-2021-evidence-based} to allow evidence to support or refute parts of the claim only. We separately mark supporting evidence from the claim's future, as the claim's veracity may have changed. We consider correctly identified counter-evidence as refuting the claim, even when it requires the source guarantee.
\section{Experiments on \textsc{MultiFC}}
\subsection{Training details}
\label{app:multifc-analysis:experiment-setup}

For our experiments we use \texttt{bert-base-uncased} as provided by \citet{wolf-etal-2020-transformers}.
We represent each evidence snippet $e$ as the title, the text body, or the concatenation of both (depending on the experiment). We concatenate all evidence snippets $e_i$, separated by a semicolon ($e_1$; $e_2$; ...; $e_n$).
We input the concatenation of the claim $c$ and the concatenated evidence, separated by \texttt{[SEP]} token, and truncated after 512 tokens:
{
\texttt{[CLS]} $c$ \texttt{[SEP]} $e_1$; $e_2$; ...; $e_n$ \texttt{[SEP]}.
}
We predict the label via a linear layer on the \textsc{[CLS]} token.
We train all models for \texttt{5} epochs with a learning rate of \texttt{2e-5} and a batch size of \texttt{16}. We select the model with the highest F1 score on the development dataset, evaluated after each epoch. We keep the default parameters for all other values. We always train and evaluate three models using the seeds \texttt{(1,2,3)}. We did not fine-tune any hyperparameter.
We provide code for reproduction. We run our experiments on a DGX A100.

\subsection{Performance per Label}
\label{app:multifc-analysis:performance-by-label}
\begin{table*}[]
\small
    \centering
    \begin{tabular}{l | c c c | c c  c| c c}
    \toprule
    & \multicolumn{3}{c|}{\textit{Leaked}} & \multicolumn{3}{c|}{\textit{Unleaked}} & \multicolumn{2}{c}{\textit{Difference}} \\
    \textbf{Gold Label} & \textbf{Ev.-Only} & \textbf{Full} & \textbf{\# Pairs} & \textbf{Ev.-Only} & \textbf{Full} & \textbf{\# Pairs} &  \textbf{$\Delta$ Ev.-Only} & \textbf{$\Delta$ Full} \\
    \toprule

\textbf{true} & 33.3 \tiny{$\pm$1.2} & 33.6 \tiny{$\pm$3.9} & 39  & 45.0 \tiny{$\pm$0.8} & 44.9 \tiny{$\pm$1.7} & 93  & -11.6 \tiny{$\pm$2.0} & -11.3 \tiny{$\pm$2.5}\\
\textbf{mostly true} & 0.0 \tiny{$\pm$0.0} & 0.0 \tiny{$\pm$0.0} & 10  & 0.0 \tiny{$\pm$0.0} & 0.0 \tiny{$\pm$0.0} & 19  & 0.0 \tiny{$\pm$0.0} & 0.0 \tiny{$\pm$0.0}\\
\textbf{mixture} & 18.0 \tiny{$\pm$3.8} & 22.4 \tiny{$\pm$3.0} & 44  & 28.6 \tiny{$\pm$2.3} & 32.8 \tiny{$\pm$1.0} & 81  & -10.6 \tiny{$\pm$5.9} & -10.4 \tiny{$\pm$2.5}\\
\textbf{mostly false} & 6.2 \tiny{$\pm$5.5} & 10.8 \tiny{$\pm$4.0} & 36  & 4.4 \tiny{$\pm$3.9} & 9.4 \tiny{$\pm$7.2} & 40  & +1.9 \tiny{$\pm$8.3} & +1.5 \tiny{$\pm$4.1}\\
\textbf{false} & 85.8 \tiny{$\pm$0.4} & 86.2 \tiny{$\pm$0.3} & 353  & 73.2 \tiny{$\pm$0.9} & 77.7 \tiny{$\pm$2.0} & 299  & +12.5 \tiny{$\pm$1.1} & +8.5 \tiny{$\pm$1.7}\\

     \bottomrule
     
    \end{tabular}
    \caption{F1-score on \textbf{Snopes} (via \textsc{MulitFC}) using the evidence-only model using solely evidence (title \& text) snippets, and the full model. We report the F1-score for each label on both splits (leaked and unleaked evidence), and their difference.}
    \label{tab:multifc-experiments-label-f1-snes}
\end{table*} 
\begin{table*}[]
\small
    \centering
    \begin{tabular}{l | c c c | c c  c| c c}
    \toprule
    & \multicolumn{3}{c|}{\textit{Leaked}} & \multicolumn{3}{c|}{\textit{Unleaked}} & \multicolumn{2}{c}{\textit{Difference}} \\
    \textbf{Gold Label} & \textbf{Ev.-Only} & \textbf{Full} & \textbf{\# Pairs} & \textbf{Ev.-Only} & \textbf{Full} & \textbf{\# Pairs} &  \textbf{$\Delta$ Ev.-Only} & \textbf{$\Delta$ Full} \\
    \toprule

\textbf{True} & 57.9 \tiny{$\pm$1.2} & 58.3 \tiny{$\pm$1.0} & 288  & 29.0 \tiny{$\pm$1.4} & 27.3 \tiny{$\pm$3.7} & 113  & +29.0 \tiny{$\pm$1.0} & +31.0 \tiny{$\pm$3.7}\\
\textbf{Mostly True} & 59.0 \tiny{$\pm$0.7} & 58.1 \tiny{$\pm$0.2} & 387  & 20.5 \tiny{$\pm$1.9} & 25.8 \tiny{$\pm$2.4} & 123  & +38.5 \tiny{$\pm$2.6} & +32.3 \tiny{$\pm$2.5}\\
\textbf{Half-True} & 58.3 \tiny{$\pm$1.3} & 58.4 \tiny{$\pm$0.3} & 405  & 28.6 \tiny{$\pm$2.4} & 30.9 \tiny{$\pm$2.6} & 132  & +29.8 \tiny{$\pm$3.6} & +27.5 \tiny{$\pm$2.3}\\
\textbf{Mostly False} & 56.2 \tiny{$\pm$1.2} & 56.8 \tiny{$\pm$0.4} & 364  & 18.2 \tiny{$\pm$6.2} & 17.1 \tiny{$\pm$0.7} & 97  & +38.0 \tiny{$\pm$7.1} & +39.7 \tiny{$\pm$0.3}\\
\textbf{False} & 58.4 \tiny{$\pm$0.9} & 58.8 \tiny{$\pm$1.1} & 419  & 19.0 \tiny{$\pm$3.0} & 26.4 \tiny{$\pm$0.9} & 102  & +39.5 \tiny{$\pm$2.8} & +32.4 \tiny{$\pm$0.9}\\
\textbf{Pants on Fire!} & 68.1 \tiny{$\pm$0.6} & 68.0 \tiny{$\pm$0.4} & 248  & 17.6 \tiny{$\pm$3.7} & 25.9 \tiny{$\pm$2.2} & 39  & +50.5 \tiny{$\pm$3.1} & +42.2 \tiny{$\pm$1.9}\\

     \bottomrule
     
    \end{tabular}
    \caption{F1-score on \textbf{PolitiFact} (via \textsc{MulitFC}) using the evidence-only model using solely evidence (title \& text) snippets, and the full model. We report the F1-score for each label on both splits (leaked and unleaked evidence) and their difference.}
    \label{tab:multifc-experiments-label-f1-pomt}
\end{table*} 
We show the F1 score for each label and both datasets in Table~\ref{tab:multifc-experiments-label-f1-snes} (Snopes) and Table~\ref{tab:multifc-experiments-label-f1-pomt} (PolitiFact).
The dataset of Snopes is highly imbalanced towards the majority class ``false''. The class imbalance is amplified within the leaked subset. 
The evidence-only model benefits from leaked evidence for (leaning) false claims. 
The performance drops on samples with evidence exhibiting leaked characteristics when the gold veracity tends towards true.
The performance on the unleaked subset is slightly more balanced when comparing different labels. The majority label ``false'' is still dominating. Yet, the more balanced predictions across all labels yield a higher F1-macro score.
On PolitiFact, a majority of claims across all labels contain leaked evidence. Models learn to exploit it for all labels. The best performance gain can be seen over claims misinformation claims.  Yet, the more balanced predictions across all labels yield a higher F1-macro score.

\subsection{Evaluation on Identical Claims with Different Evidence}
\label{app:multifc-analysis:prformance-by-only-leaked-unleaked}
\begin{table*}[]
\small
    \centering
    \begin{tabular}{l | c c c | c c c| c }
    \toprule
    & \multicolumn{3}{c|}{\textit{Leaked}} & \multicolumn{3}{c|}{\textit{Unleaked}} &  \\
    \textbf{Gold Label} & \textbf{Precision} & \textbf{Recall} & \textbf{F1} & \textbf{Precision} & \textbf{Recall} & \textbf{F1} &  \textbf{Support} \\
    \toprule

    \textbf{true} & \textbf{31.5} & 14.0 & 19.0 & 17.0 & \textbf{33.3} & \textbf{22.5} & 38 \\
\textbf{mostly true} & 0.0 & 0.0 & 0.0 & 0.0 & 0.0 & 0.0 & 10 \\
\textbf{mixture} & \textbf{28.1} & 23.5 & 25.4 & 19.8 & \textbf{37.1} & \textbf{25.8} & 44 \\
\textbf{mostly false} & 0.0 & 0.0 & 0.0 & \textbf{5.9} & \textbf{2.8} & \textbf{3.8} & 36 \\
\textbf{false} & 77.3 & \textbf{93.5} & \textbf{84.6} & \textbf{82.4} & 73.9 & 77.9 & 353 \\
\midrule
\textbf{Accuracy} & \multicolumn{3}{c|}{\textbf{71.9}} & \multicolumn{3}{c|}{60.5} & 481 \\
\textbf{F1-micro} & \multicolumn{3}{c|}{\textbf{66.0}} & \multicolumn{3}{c|}{61.6} & 481 \\
\textbf{F1-macro} & \multicolumn{3}{c|}{25.8} & \multicolumn{3}{c|}{\textbf{26.0}} & 481 \\

     \bottomrule
     
    \end{tabular}
    \caption{Precision recall and F1 of the evidence-only BERT model based on all claims of the \textbf{Snopes} dataset that contain leaked and unleaked evidence snippets.}
    \label{tab:multifc-experiments-same-claim-different-evidence-snopes}
\end{table*} 
\begin{table*}[]
\small
    \centering
    \begin{tabular}{l | c c c | c c c| c }
    \toprule
    & \multicolumn{3}{c|}{\textit{Leaked}} & \multicolumn{3}{c|}{\textit{Unleaked}} &  \\
    \textbf{Gold Label} & \textbf{Precision} & \textbf{Recall} & \textbf{F1} & \textbf{Precision} & \textbf{Recall} & \textbf{F1} &  \textbf{Support} \\
    \toprule

\textbf{true} & \textbf{58.3} & \textbf{53.9} & \textbf{56.0} & 21.1 & 23.8 & 22.3 & 288 \\
\textbf{mostly true} & \textbf{61.4} & \textbf{54.5} & \textbf{57.0} & 26.7 & 26.6 & 26.5 & 385 \\
\textbf{half-true} & \textbf{46.7} & \textbf{61.6} & \textbf{52.8} & 24.2 & 32.3 & 27.6 & 404 \\
\textbf{mostly false} & \textbf{59.1} & \textbf{55.4} & \textbf{57.1} & 21.8 & 18.4 & 19.7 & 360 \\
\textbf{false} & \textbf{60.9} & \textbf{58.1} & \textbf{59.4} & 26.4 & 28.5 & 27.4 & 419 \\
\textbf{pants on fire!} & \textbf{75.0} & \textbf{63.0} & \textbf{68.5} & 57.4 & 22.3 & 32.1 & 247 \\
\midrule
\textbf{Accuracy} & \multicolumn{3}{c|}{\textbf{57.6}} & \multicolumn{3}{c|}{25.8} & 2103 \\
\textbf{F1-micro} & \multicolumn{3}{c|}{\textbf{57.9}} & \multicolumn{3}{c|}{25.8} & 2103 \\
\textbf{F1-macro} & \multicolumn{3}{c|}{\textbf{58.5}} & \multicolumn{3}{c|}{25.9} & 2103 \\

     \bottomrule
     
    \end{tabular}
    \caption{Precision recall and F1 of the evidence-only BERT model based on all claims of the \textbf{PolitiFact} dataset that contain leaked and unleaked evidence snippets.}
    \label{tab:multifc-experiments-same-claim-different-evidence-politifact}
\end{table*} 
To avoid the label distribution to distort the impact of leaked evidence, we evaluate the trained model only on claims that contain leaked \textit{and} unleaked evidence snippets. We separately measure the performance of the evidence-only models on this subset, when only concatenating leaked evidence to the claim, and when only concatenating unleaked evidence to the same claim.
The overall metrics on Snopes (Table~\ref{tab:multifc-experiments-same-claim-different-evidence-snopes}) do not change much. The performance on the individual labels differs (even not always for the better) when comparing the prediction with leaked or unleaked evidence. We find that leaked evidence snippets are strong indicators for the model to predict the label ``false''. This comes at the cost of a lower recall across all other labels. 
On PolitiFact evidence snippets show great improvements and double almost every metric (Table~\ref{tab:multifc-experiments-same-claim-different-evidence-politifact}).

\subsection{Comparison with a Claim-Only Baseline}
\begin{table*}[]
\small
    \centering
    \begin{tabular}{l | c c c c| c c c c }
    \toprule
    & \multicolumn{4}{c|}{\textit{Snopes}} & \multicolumn{4}{c}{\textit{PolitiFact}} \\
    \textbf{Gold Label} & \textbf{Precision} & \textbf{Recall} & \textbf{F1} & \textbf{Support} & \textbf{Precision} & \textbf{Recall} & \textbf{F1} &  \textbf{Support} \\
    \toprule

\textbf{true} & 41.8 & 33.3 & 36.9 & 38 & 25.6 & 25.6 & 25.5 &  288 \\
\textbf{mostly true} & 0.0 & 0.0 & 0.0 & 10 & 29.8 & 34.1 & 31.7 &  385 \\
\textbf{half-true / mixture} & 22.6 & 28.0 & 24.9 & 44 & 26.7 & 35.1 & 30.3 &  404 \\
\textbf{mostly false} & 7.3 & 1.9 & 2.9 & 36 & 23.0 & 21.5 & 22.2 &  360 \\
\textbf{false} & 80.7 & 88.7 & 84.5 & 353 & 28.2 & 23.9 & 25.9 & 419 \\
\textbf{pants on fire!} & -- & -- & -- &  -- & 54.7 & 33.3 & 41.3 & 247 \\
\toprule
\textbf{Accuracy} & \multicolumn{3}{c}{70.4} & 481 & \multicolumn{3}{c}{28.8} & 2103 \\
\textbf{F1-micro} & \multicolumn{3}{c}{67.4} & 481 & \multicolumn{3}{c}{28.9} & 2103 \\
\textbf{F1-macro} & \multicolumn{3}{c}{29.8} & 481 & \multicolumn{3}{c}{29.5} & 2103 \\

     \bottomrule
     
    \end{tabular}
    \caption{Precision recall and F1 of the \textit{claim-only} BERT model based on all claims containing leaked and unleaked evidence.}
    \label{tab:multifc-experiments-same-claim-different-evidence-claim-only-both}
\end{table*}
We show the detailed results of the \textit{claim-only} baseline for all claims that contain leaked and unleaked evidence snippets in Table~\ref{tab:multifc-experiments-same-claim-different-evidence-claim-only-both}. All these samples are considered leaked, as they contain (amongst others) leaked evidence. The results align with our previous observations: Models trained on the Snopes subset rely on the majority class ``false'', yielding an overall high performance. 
We compare the results of the claim-only model (which by default cannot benefit from leaked evidence) with the results of the evidence-only model (compare Table~\ref{tab:multifc-experiments-same-claim-different-evidence-snopes}) on the same subset.
The claim-only model outperforms the evidence-only model on unleaked evidence. When the evidence-only model sees leaked evidence, it either performs on par (F1-micro, accuracy) or worse  (F1-macro) than the claim-only model. Here, the claim-only model is slightly less biased towards predicting ``false'', which improves the performance on the remaining labels. Both results indicate the reliance of the model on leaked evidence (even if not for the better).
On the PolitiFact subset, the claim-only baseline performs well behind the evidence-only baseline (compare Table~\ref{tab:multifc-experiments-same-claim-different-evidence-politifact}).

\end{document}